# Makespan Optimal Solving of Cooperative Path-Finding via Reductions to Propositional Satisfiability

## Pavel Surynek

*National Institute of Advanced Industrial Science and Technology (AIST)*
*AIRC, 2-3-26, Aomi, Koto-ku, Tokyo 135-0064, Japan*

*Charles University, Faculty of Mathematics and Physics*
*Malostranské náměstí 25, 118 00 Praha 1, Czechia*

pavel.surynek@aist.go.jp

*Abstract.* The problem of makespan optimal solving of cooperative path finding (CPF) is addressed in this paper. The task in CPF is to relocate a group of agents in a non-colliding way so that each agent eventually reaches its goal location from the given initial location. The abstraction adopted in this work assumes that agents are discrete items moving in an undirected graph by traversing edges. Makespan optimal solving of CPF means to generate solutions that are as short as possible in terms of the total number of time steps required for the execution of the solution.

We show that reducing CPF to propositional satisfiability (SAT) represents a viable option for obtaining makespan optimal solutions. Several encodings of CPF into propositional formulae are suggested and experimentally evaluated. The evaluation indicates that SAT based CPF solving outperforms other makespan optimal methods significantly in highly constrained situations (environments that are densely occupied by agents).

## 1. Introduction and Motivation

*Cooperative path-finding* - CPF [14, 23, 25] (also known as *multi-agent path finding* - MAPF [21, 22, 37, 38] or as *multi-robot path planning* - MRPP [18, 19] or as *pebble motion on a graph* - PMG [14, 16]) is an abstraction for many real-file tasks where the goal is to relocate some objects that spatially interacts with each other. In case of CPF, we are speaking about mobile *agents* (or robots) that can be moved in a certain environment. Each agent starts at a given initial position in the environment and it is assigned a unique goal position to which it has to relocate. The problem consists in finding a spatial-temporal path for each agent by which the agent can relocate itself from its initial position to the given goal without colliding with other agents (that are simultaneously trying to reach their goals as well).

A graph theoretical abstraction, where the environment in which agents are moving is modeled as an undirected graph, is often adopted [18, 20]. Agents are represented as discrete items placed in vertices of the graph in this abstraction. Space occupancy imposed by presence of agents is modeled by the requirement that at most one agent resides in each vertex.

Movements of agents are also greatly simplified in the abstraction. An agent can instantaneously move to a neighboring vertex assumed that the target vertex is unoccupied and no other agent is trying to enter the same target vertex simultaneously. Note that various versions of the problem may have different conditions on movements - sometimes it is for instance allowed to move agents in a train like manner [28] or even rotate agents around cycle without any unoccupied vertex in the cycle [39].

---

[1] This is a preprint submitted to ArXiv on October 18, 2016.

There are many practical motivations for CPF ranging from unit navigation in computer games [24] to item relocation in automated storage (see KIVA robots [13]). Interesting motivations can be also found in traffic where problems like vessel avoidance at sea are of great practical importance [12]. An analogical challenge appears in the air where availability of drones implies need for developing cooperative air traffic control mechanisms [15].

We suggest to solve CPF via reducing it to *propositional satisfiability* (SAT) [7]. Particularly we are dealing with so-called *makespan optimal* solving of CPF [23, 29], which means to find a solution of a makespan as short as possible. The makespan of a solution is the number of steps necessary to execute all the moves of the solution. In other words, it is the length of the longest path from paths traveled by individual agents. It is known that finding makespan optimal solutions to CPF is a difficult problem, namely it is NP-hard [16, 32, 39]. Hence reducing the makespan optimal CPF to SAT is justified as both problems are at the same level in terms of the complexity. Moreover, the reduction allows exploiting the power of modern SAT solvers [2, 3] in CPF solving. The question however is the design of an encoding of the CPF problem into propositional formula. Several encodings of CPF into propositional formulae are introduced in this paper. They are based on a so-called *time expansion* of the graph that models the environment [11, 26] so that the formula can represent all the possible arrangements of agents at all the time steps up to the given final time step. All the encodings are thoroughly experimentally evaluated with each other and also with alternative techniques for makespan optimal CPF solving.

## 2. Context of Related Works

The approach to solve CPF by reducing it to SAT has multiple alternatives. There exist algorithms based on search that find makespan optimal or near optimal solutions. The seminal work in this category is represented by Silver's WHCA* algorithm [20] which is a variant of A* search where cooperation among agents is incorporated. Recent contributions include OD+ID [23], which is a combination of A* and powerful agent independence detection heuristics, and ICTS [21] which employs the concept of increasing cost tree (instead of makespan, the total cost of solution is optimized). Other approaches resolve conflicts among robot trajectories when avoidance is necessary [5, 8, 34].

Fast polynomial time algorithms for generating makespan suboptimal solutions include PUSH-AND-ROTATE [37, 38] and other algorithms [28]. The drawback of these algorithms is that their solutions are dramatically far from the optimum.

Translation of CPF to a different formalism, namely to answer set programming (ASP), has been suggested in [9]. Integer programming (IP) as the target formalism has been also used [39]. The choice of SAT as the target formalism is very common in domain independent planning where the idea of time expansion [10, 11] and its reductions [4, 35] are studied.

## 3. Background

An arbitrary **undirected graph** can model the environment where agents are moving. Let $G = (V, E)$ be such a graph where $V = \{v_1, v_2, ..., v_n\}$ is a finite set of vertices and $E \subseteq \binom{V}{2}$ is a set of edges. The configuration of agents in the environment is modeled by assigning them vertices of the graph. Let $A = \{a_1, a_2, ..., a_\mu\}$ be a finite set of *agents*. Then, a configuration of agents in vertices of graph $G$ will be fully described by a *location* function $\alpha: A \rightarrow V$; the interpretation is that an agent $a \in A$ is located in a vertex $\alpha(a)$. At most **one agent** can be located in a vertex; that is $\alpha$ is a uniquely invertible function. A generalized inverse of $\alpha$ denoted as $\alpha^{-1}: V \rightarrow A \cup \{\bot\}$ will provide us an agent located in a given vertex or $\bot$ if the vertex is empty.

**Definition 1** (COOPERATIVE PATH FINDING). An instance of *cooperative path-finding* problem (CPF) is a quadruple $\Sigma = [G = (V, E), A, \alpha_0, \alpha_+]$ where location functions $\alpha_0$ and $\alpha_+$ define the initial and the goal configurations of a set of agents $A$ in $G$ respectively. □

The dynamicity of the model assumes a discrete time divided into time steps. A configuration $\alpha_i$ at the $i$-th time step can be transformed by a transition action which instantaneously moves agents in the non-colliding way to form a new configuration $\alpha_{i+1}$. The resulting configuration $\alpha_{i+1}$ must satisfy the following *validity conditions*:

- $\forall a \in A$ either $\alpha_i(a) = \alpha_{i+1}(a)$ or $\{\alpha_i(a), \alpha_{i+1}(a)\} \in E$ holds (1)
  (agents move along edges or not move at all),

- $\forall a \in A \;\; \alpha_i(a) \neq \alpha_{i+1}(a) \Rightarrow \alpha_i^{-1}(\alpha_{i+1}(a)) = \bot$ (2)
  (agents move to vacant vertices only), and

- $\forall a, b \in A \;\; a \neq b \Rightarrow \alpha_{i+1}(a) \neq \alpha_{i+1}(b)$ (3)
  (no two agents enter the same target/unique invertibility of resulting arrangement).

The task in cooperative path finding is to transform $\alpha_0$ using above valid transitions to $\alpha^+$. An illustration of CPF and its solution is depicted in Figure 1.

**Definition 2** (SOLUTION, MAKESPAN). A *solution* of a *makespan* $\eta$ to a cooperative path finding instance $\Sigma = [G, A, \alpha_0, \alpha^+]$ is a sequence of arrangements $\vec{s} = [\alpha_0, \alpha_1, \alpha_2, ..., \alpha_\eta]$ where $\alpha_\eta = \alpha^+$ and $\alpha_{i+1}$ is a result of valid transformation of $\alpha_i$ for every $i = 1,2, ..., \eta - 1$. □

The number $|\vec{s}| = \eta$ is a *makespan* of solution $\vec{s}$. It is often a question whether there exists a solution of $\Sigma$ of the given makespan $\eta \in \mathbb{N}$. This is known as a *decision variant of CPF*. It is known that the decision variant of CPF is NP-complete, hence finding makespan optimal solution to CPF is NP-hard [16]. Note that due to no-ops introduced in valid transitions, it is equivalent to ask whether there is a solution of exactly the given makespan ant to ask whether there is a solution of at most given makespan.

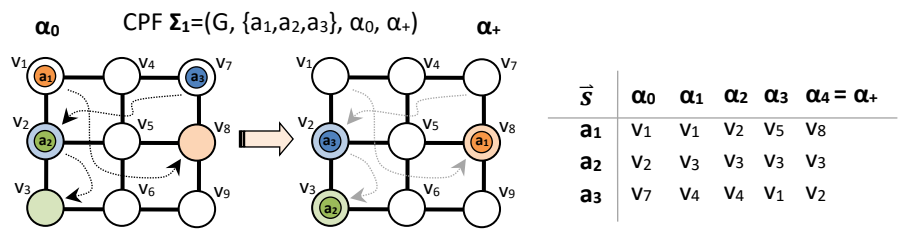

**Figure 1.** *An example of cooperative path-finding problem (CPF).* Three agents $a_1$, $a_2$, and $a_3$ need to relocate from their initial positions represented by $\alpha_0$ to goal positions represented by $\alpha_+$. A solution of makespan 4 is shown.

## 4. Solving CPF Optimally through Propositional Satisfiability

The question we are addressing is how to obtain *makespan optimal solutions* of CPFs in some practical manner. The approach we are suggesting here employs *propositional satisfiability* (SAT) [1] solving as the key technology. Note that the decision variant of CPF is in NP, hence it can be reduced to propositional satisfiability. That is, a propositional formula $F(\Sigma, \eta)$ such that it is satisfiable if and only if a given CPF $\Sigma$ with makespan $\eta$ is solvable can be constructed. Being able to construct such a formula $F(\Sigma, \eta)$ one can obtain the optimal makespan for the given CPF $\Sigma$ by asking multiple queries whether formula $F(\Sigma, \eta)$ is satisfiable with different makespan bounds $\eta$.

Various strategies of choice of makespan bounds for queries exist for getting the optimal makespan. The simplest and efficient one at the same time is to try sequentially makespan $\eta = 1, 2, ...$ until $\eta$ equal to the optimal makespan is reached. This strategy will be further referred as *sequential increasing*. The sequential increasing strategy is also used in domain independent planners such as SATPLAN [11], SASE [10] and others. Pseudo-code of the strategy is listed as Algorithm 1.

The focus here is on SAT encoding while querying strategies are out of scope of the paper; though let us mention that in depth study of querying strategies is given in [17]. There is a great potential in querying strategies as they can bring speedup of planning process in orders of magnitude, especially when combined with parallel processing.

The important property of propositional encoding $F(\Sigma, \eta)$ is that a solution of CPF $\Sigma$ of makespan $\eta$ can be unambiguously extracted from satisfying valuation of $F(\Sigma, \eta)$ (otherwise, equivalence between solvability of CPF $\Sigma$ bounded by $\eta$ and solvability of $F(\Sigma, \eta)$ could be trivially established by setting $F(\Sigma, \eta) \equiv TRUE$ in case $\Sigma$ is solvable in $\eta$ time steps and $F(\Sigma, \eta) \equiv FALSE$ otherwise).

Note that the solving process represented by Algorithm 1 is incomplete, as it does not terminate when the input instance is unsolvable. Nevertheless, the solving process can be easily made complete by checking instance solvability prior to SAT-based optimization by some fast polynomial time algorithm such as those described in [14, 28, 38].

**Algorithm 1.** *SAT-based optimal CPF solving – sequential increasing strategy.* The algorithm sequentially finds the smallest possible makespan $\eta$ for that a given CPF $\Sigma = (G, A, \alpha_0, \alpha_+)$ is solvable. A question whether a solution of CPF $\Sigma$ exists is constructed with respect to increasing makespans and submitted to a SAT solver.

    **input:**    $\Sigma$ – a CPF instance
    **output:**  a pair consisting of the optimal makespan and corresponding optimal solution

**function** *Find-Optimal-Solution-Sequentially* ($\Sigma = (G, A, \alpha_0, \alpha_+)$): **pair**
1:   $\eta \leftarrow 1$
2:   **loop**
3:       $F(\Sigma, \eta) \leftarrow$ *Encode-CPF-as-SAT* $(\Sigma, \eta)$
4:       **if** *Solve-SAT* $(F(\Sigma, \eta))$ **then**
5:           $s \leftarrow$ *Extract-Solution-from-Valuation*$(F(\Sigma, \eta))$
6:           **return** $(\eta, s)$
7:       $\eta \leftarrow \eta + 1$
8:   **return** $(\infty, \emptyset)$

    The important advantage of solving CPF as SAT is that there exist many powerful solvers for SAT [2, 3] implementing numerous advanced techniques such as intelligent search space pruning and learning. The spectrum of these techniques is so rich and so well engineered in modern SAT solvers that it is almost impossible to reach the equal level of advancement in solving CPF by own dedicated solver. Nevertheless, all the well-engineered techniques implemented in SAT solvers can be employed in CPF solving if it is translated to SAT. Note, that the effect of SAT solving techniques is indirect in CPF solving as it is mediated through the translation. Hence, the design of the encoding of CPF as SAT should take into consideration the way in which SAT solvers operate.

### 4.1. Time Expansion Graphs

The *trajectory* of an agent in time over $G$ is not necessarily *simple* in general case (that is, a single vertex can be visited multiple times). In a propositional representation of such kind of trajectory, it is difficult to fix the number of variables. Therefore, a graph derived from $G$ by expanding it over time, where the trajectory of each agent will correspond to a simple path in this graph, will be used (a simple path visits each vertex of the graph at most once). The graph of required properties is introduced in the following definition and illustrated in Figure 3.

**Definition 3** (TIME EXPANSION GRAPH - $\text{Exp}_T(G, \eta)$). Let $G = (V, E)$ be an undirected graph and $\eta \in \mathbb{N}$. A *time expansion graph* with $\eta + 1$ time layers (indexed from 0 to $\eta$) associated with $G$ is a directed graph $\text{Exp}_T(G, \eta) = (V \times \{0, 1, \dots, \eta\}, E')$ where $E' = \{([u, l], [v, l+1]) \mid \{u, v\} \in E; l = 0, 1, \dots, \eta - 1\} \cup \{([v, l], [v, l+1]) \mid v \in V; l = 0, 1, \dots, \eta - 1\}$. □

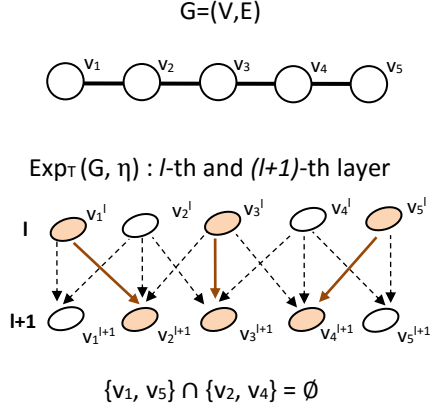

**Figure 2.** *An illustration of non-overlapping vertex disjoint paths*. Parts of three non-overlapping paths between time layers $l$ and $l+1$ of $\text{Exp}_T(G,\eta)$ are shown.

Notation $u^l$ will be sometimes used instead of $[u, l]$ in figures. The search for a solution of CPF with makespan bound $\eta$ can be viewed as the search for a collection of so-called *non-overlapping vertex disjoint paths* in the corresponding time expansion graph consisting of $\eta$ layers $\text{Exp}_T(G,\eta)$. This is also the reason why the number of time layers in time expansion graphs and the makespan bound in CPF use the same notation with $\eta$. Non-overlapping vertex disjoint paths must have disjoint set of endpoints of non-trivial edges in consecutive time layers of $\text{Exp}_T(G,\eta)$ as described in the following definition.

**Definition 4** (NON-OVERLAPPING VERTEX DISJOINT PATHS IN $\text{Exp}_T(G,\eta)$). A collection of paths $\Pi = \{\pi_1, \pi_2, \ldots, \pi_\mu\}$ in $\text{Exp}_T(G,\eta)$ so that $\pi_i$ connects $[x_i, 0]$ with $[y_i, \eta]$ with $x_i, y_i \in V$ for $i = 1, 2, \ldots, \mu$ is called to be *non-overlapping vertex disjoint* if and only if $\pi_i \cap \pi_j = \emptyset$ for any two $i, j \in \{1, 2, \ldots, \mu\}$ with $i \neq j$ and $\{\pi_i[l, 2] \mid \pi_i[l, 2] \neq \pi_i[l+1, 2] \wedge i = 1, 2, \ldots, \mu\} \cap \{\pi_i[l+1, 2] \mid \pi_i[l, 2] \neq \pi_i[l+1, 2] \wedge i = 1, 2, \ldots, \mu\}$[1] for $l = 0, 1, \ldots, \eta - 1$. □

Non-overlapping vertex disjoint paths between two consecutive time layers of $\text{Exp}_T(G,\eta)$ are shown in Figure 2. The correspondence between existence of a solution to CPF and non-overlapping vertex disjoint paths is established in the next proposition.

**Proposition 1** (*NON-OVERLAPPING VERTEX DISJOINT PATHS IN $\text{Exp}_T$*). *A solution of makespan $\eta \in \mathbb{N}$ of a CPF $\Sigma = (G, A, \alpha_0, \alpha_+)$ with $A = \{a_1, a_2, \ldots, a_\mu\}$ exists if and only if there exist a set $\Pi = \{\pi_1, \pi_2, \ldots, \pi_\mu\}$ of non-overlapping vertex disjoint paths in $\text{Exp}_T(G,\eta)$ so that $\pi_i$ connects $[\alpha_0(a_i), 0]$ with $[\alpha_+(a_i), \eta]$ for $i = 1, 2, \ldots, \mu$.* ∎

**Proof.** Assume that a solution $\vec{s} = [\alpha_0, \alpha_1, \alpha_2, \ldots, \alpha_\eta]$ of makespan $\eta$ of given CPF $\Sigma$ exists. Then vertex disjoint paths $\pi_1, \pi_2, \ldots, \pi_\mu$ in $\text{Exp}_T(G,\eta)$ can be constructed from $\vec{s}$. Path $\pi_i$ will correspond to the trajectory of agent $a_i$; that is, $\pi_i = ([\alpha_0(a_i), 0], [\alpha_1(a_i), 1], \ldots, [\alpha_\eta(a_i), \eta])$. The path constructed in this way is a correct path in $\text{Exp}_T(G,\eta)$, since $\{\alpha_l(a_i), \alpha_{l+1}(a_i)\} \in E$ or $\alpha_l(a_i) = \alpha_{l+1}(a_i)$ for $l = 0, 1, \ldots, \eta - 1$; that is, $([\alpha_l(a_i), l], [\alpha_{l+1}(a_i), l+1]) \in E'$ holds by construction of $\text{Exp}_T(G,\eta)$. Obviously $\pi_i$ connects $[\alpha_0(a_i), 0]$ with $[\alpha_+(a_i) = \alpha_\eta(a_i), \eta]$ in $\text{Exp}_T(G,\eta)$. It remains to check that no two constructed paths intersect and that paths are non-overlapping. Validity condition (3) ensures that no two path share a common vertex since otherwise agents would collide. Validity conditions (1) and (2) together ensure that overlapping between set of endpoints of edges of paths between consecutive time layers happens only with trivial edges – that is, edges that continues into the same vertex in the next time layer.

---

[1] The notation $\pi_i[l, 2]$ refers to the second component of the $l$-th element of $\pi_i$.

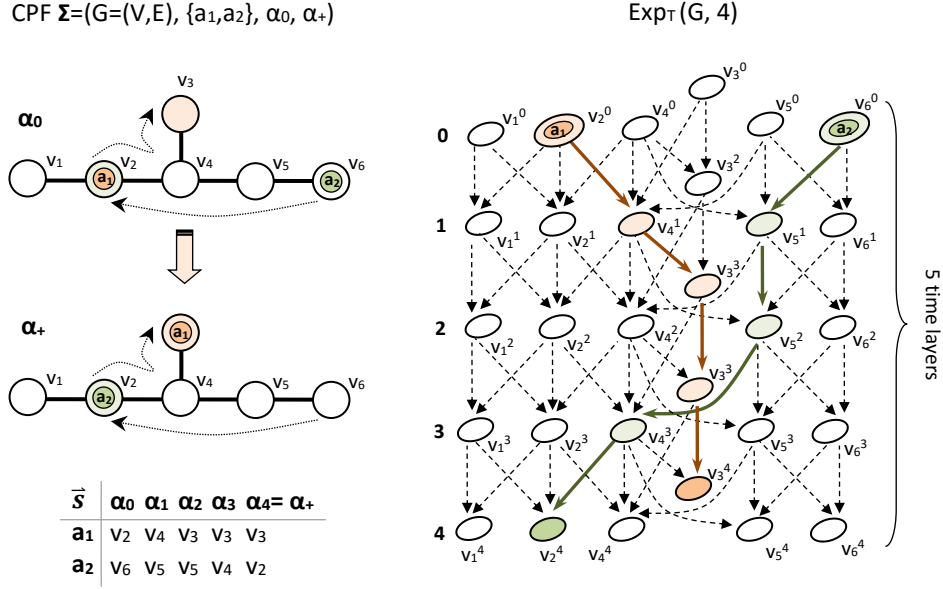

**Figure 3.** *An example of CPF and its time expansion graph.* A time expansion graph $\text{Exp}_T(G, 4)$ consisting of 5 time layers is build for a given CPF Σ. Solving Σ in 5 time steps can be represented as searching for a collection of non-overlapping vertex disjoint paths connecting the initial positions agents in the first layer with their goal positions in the last layer of $\text{Exp}_T(G, 4)$.

Let us show the opposite implication. Assume that non-overlapping vertex disjoint paths $\pi_1, \pi_2, \ldots, \pi_\mu$ in $\text{Exp}_T(G, \eta)$ exist. We will construct a solution of CPF Σ of makespan $\eta$. Assume that Let $\pi_i = ([u_0, 0], [u_1, 1], [u_2, 2], \ldots, [u_\eta, \eta])$, $u_l \in V$ for $l = 0, 1, \ldots, \eta$ where $u_0 = \alpha_0(a_i)$ and $u_\eta = \alpha_+(a_i)$. The trajectory of agent $a_i$ is set as follows: $\alpha_0(a_i) = u_0, \alpha_1(a_i) = u_1, \alpha_2(a_i) = u_2, \ldots, \alpha_\eta(a_i) = u_\eta$. It can be easily verified that validity conditions (1) – (3) are satisfied by such a construction. Paths are vertex disjoint, so agents do not collide by following them – condition (2) is satisfied. As paths do not overlap agents either stay in a vertex or move into a vertex that was not occupied in the previous step. Altogether, validity conditions (1) – (3) are satisfied. ∎

### 4.2. Propositional Encodings Based on Time Expansion Graphs

The concept of time expansion graph represents an important step towards the design of a propositional formula that is satisfiable if and only if the given CPF has a solution of a given makespan. Moreover, we require such a formula where a corresponding CPF solution can be extracted from its satisfying valuation. Time expansion graph can be used as a basis for such a formula as it can capture all the arrangements of agents over the graph modeling the environment at all the time steps up to the given final step.

### 4.2.1. INVERSE Propositional Encoding

Let $\deg_G(v)$ denote the degree of vertex $v$ in $G$; that is, $\deg_G(v)$ is the number of edges from $E$ incident with $v$. It is further assumed that neighbors of each vertex $v$ in $G$ are assigned ordering numbers by a one-to-one assignment $\sigma_v : \{u | \{v, u\} \in E\} \to \{1, 2, \ldots, \deg_G(v)\}$ (that is, for each neighbor $u$ of $v$ we are told that it is a $\sigma_v(u)$-th neighbor). An inverse $\sigma_v^{-1}$ is naturally defined (that is, $\sigma_v^{-1}(i)$ returns $i$-th neighbor of $v$ for $i \in \{1, 2, \ldots, \deg_G(v)\}$).

The following definition introduces the INVERSE encoding over finite domain state variables that will be further encoded into bit-vectors using the standard *binary encoding*.

**Definition 5** (INVESE ENCODING – $F_{INV}(\eta, \Sigma)$). Assume that a CPF $\Sigma = [G, A, \alpha_0, \alpha_+]$ with $G = (V, E)$ is given. An *INVERSE encoding* for CPF $\Sigma$ consists of the following finite domain variables for each time layer $l \in \{0, 1, \ldots, \eta\}$: $\mathcal{A}_v^l \in \{0, 1, \ldots, \mu\}$ for every $v \in V$ to model agent occurrences in vertices. For time layers $l \in \{0, 1, \ldots, \eta - 1\}$ there are also finite domain variables $\mathcal{T}_v^l \in \{0, 1, \ldots, 2 \cdot \deg_G(v)\}$ for every $v \in V$ to represent agent movements. Constraints of INVERSE encoding are as follows:

- $\mathcal{T}_v^l = 0 \Rightarrow \mathcal{A}_v^{l+1} = \mathcal{A}_v^l \quad$ for every $v \in V$ and $l \in \{0, 1, \ldots, \eta - 1\}$ (4)
  (if there is no movement occurs in a vertex then the vertex hold the same agent at the next time step)

- $0 < \mathcal{T}_v^l \leq \deg_G(v) \Rightarrow \mathcal{A}_u^l = 0 \land \mathcal{A}_u^{l+1} = \mathcal{A}_v^l \land \mathcal{T}_u^l = \sigma_u(v) + \deg_G(u),$ (5)
  for every $v \in V$ and $l \in \{0, 1, \ldots, \eta - 1\}$, where $u = o_v^{-1}(\mathcal{T}_v^l)$
  (an agent leaves from $v$ to its $\mathcal{T}_v^l$-th neighbor $u$)

- $\deg_G(v) < T_v^l \leq 2 \cdot \deg_G(v) \Rightarrow \mathcal{T}_u^l = \sigma_u(v),$
  for every $v \in V$ and $l \in \{0, 1, \ldots, \eta - 1\}$, where $u = \sigma_v^{-1}(\mathcal{T}_v^l - \deg_G(v))$ (6)
  (an agent leaves arrives to $v$ from its $(\mathcal{T}_v^l - \deg_G(v))$-th neighbor $u$). □

Initial and goal arrangements will be expressed though the following constraints:

- $\mathcal{A}_u^0 = i \quad$ for $u \in V$ if there is $i \in \{1, 2, \ldots, \mu\}$ such that $\alpha_0(a_i) = u$ } Initial locations (7)
- $\mathcal{A}_u^0 = 0 \quad$ for $u \in V$ if $(\forall a \in A)\alpha_0(a) \neq u$ (8)
- $\mathcal{A}_u^\eta = i \quad$ for $u \in V$ if there is $i \in \{1, 2, \ldots, \mu\}$ such that $\alpha_+(a_i) = u$ } Goal locations (9)
- $\mathcal{A}_u^\eta = 0 \quad$ for $u \in V$ if $(\forall a \in A)\alpha_+(a) \neq u$ (10)

The resulting propositional formula in CNF, where $\mathcal{A}_v^l$ and $\mathcal{T}_v^l$ variables are replaced with bit vectors with binary encoding and constraints are replaced accordingly, will be denoted as $F_{INV}(\eta, \Sigma)$.

The meaning of $\mathcal{A}_v^l$ variables correspond to the inverse location function at time step $l$. That is, if the inverse location function at time step $l$ is $\alpha_l^{-1}$ then $\mathcal{A}_v^l = j$ iff $\alpha_l^{-1}(v) = a_j$ and $\mathcal{A}_v^l = 0$ iff $\alpha_l^{-1}(v) = \perp$. Variables $\mathcal{T}_v^l$ represent transitions of agents among vertices. Zero value is reserved for no-movement. Half of remaining values from 1 to $\deg_G(v)$ represent outgoing movements from $v$ to some neighbor indicated by $\mathcal{T}_v^l$; the other half of values represent incoming movements into $v$ from some of its neighbors indicated by $\mathcal{T}_v^l - \deg_G(v)$.

It is not straightforward to encode the above finite domain model into propositional model where finite domain state variables are replaced with *bit-vectors* (vectors of propositional variables) using *binary encoding* as we need to represent quite complex integer constraints over bit vectors. Variables $\mathcal{A}_v^l$ are modeled by a vector of $\lceil \log_2(\mu + 1) \rceil$ propositional variables where individual (propositional) bits will be accessed by a bit index $\hat{\imath} \in \{0,1,\ldots,\lceil \log_2(\mu + 1) \rceil - 1\}$ denoted as $\mathcal{A}_v^l[\hat{\imath}]$. Variables $\mathcal{T}_v^l$ are modeled by vectors of $\lceil \log_2(2 \cdot \deg_G(v) + 1) \rceil$ propositional variables. Note, that typical environments are connected only locally, which means that $\deg_G(v) \ll \mu$ typically. If the represented finite domain variable has the number of states that is different from the power of 2, then extra states are forbidden.

Constraints need to distinguish between all the $2 \cdot \deg_G(v) + 1$ states of $\mathcal{T}_v^l$ variables since over bit vectors we are able to express very simple constraints only – such as an expression that a bit vector equals to a constant. Note that over $\mathcal{A}_v^l$ variables we only need to model equality between them and equality to zero which does not distinguish between too many cases. Let $\mathbb{b}: \mathbb{N}_0 \times \mathbb{N}_0 \to \{0,1\}$ be a binary representation of positive integers where $\mathbb{b}(x, \hat{\imath})$ represents value of the $\hat{\imath}$-th bit in binary encoding of $x$; that is $x = \sum_{\hat{\imath}=0}^{b-1} \mathbb{b}(x, \hat{\imath}) \cdot 2^{\hat{\imath}}$.

Equality of a $\mathcal{T}_v^l$ variable to a given constant $c \in \{0,1,\ldots,2 \cdot \deg_G(v)\}$ will be expressed as following conjunction:

$$\mathrm{con}_=(\mathcal{T}_v^l, c) = \bigwedge_{\hat{\imath}=0}^{\lceil \log_2(2 \cdot \deg_G(v)+1) \rceil - 1} \mathrm{lit}(\mathcal{T}_v^l, c, \hat{\imath}) \tag{11}$$

$$\text{where } \mathrm{lit}(\mathcal{T}_v^l, c, \hat{\imath}) = \begin{cases} \mathcal{T}_v^l[\hat{\imath}] & \text{iff } \mathbb{b}(c, \hat{\imath}) = 1 \\ \neg \mathcal{T}_v^l[\hat{\imath}] & \text{iff } \mathbb{b}(c, \hat{\imath}) = 0 \end{cases}$$

Equality between variables $\mathcal{A}_v^l$ and $\mathcal{A}_u^{l+1}$ is expressed by the following conjunction of equivalences:

$$\mathrm{var}_=(\mathcal{A}_v^l, \mathcal{A}_u^{l+1}) = \bigwedge_{\hat{\imath}=0}^{\lceil \log_2(\mu+1) \rceil - 1} (\neg \mathcal{A}_v^l[\hat{\imath}] \vee \mathcal{A}_u^{l+1}[\hat{\imath}]) \wedge (\mathcal{A}_v^l[\hat{\imath}] \vee \neg \mathcal{A}_u^{l+1}[\hat{\imath}]) \tag{12}$$

The above elementary constructions are put together to represent constraints (4) – (6) using Tseitin's encoding [33] which introduces auxiliary propositional variables to the encoding. Auxiliary propositional variables $a_{v,l}^{\mathrm{zero}}$ representing empty vertex $v$ at time step $l$, $a_{u,v,l}^=$ representing equality between $\mathcal{A}_v^l$ and $\mathcal{A}_u^{l+1}$, and $a_{v,l,c}^{\mathrm{tran}}$ representing equality $\mathcal{T}_v^l = c$.

The connection of auxiliary variables with their exact meaning is done by the following constraints:

$$a_{v,l}^{\text{zero}} \Rightarrow \text{con}_=(\mathcal{A}_v^l, 0) \tag{13}$$

$$a_{u,v,l}^= \Rightarrow \text{var}_=(\mathcal{A}_v^l, \mathcal{A}_u^{l+1}) \tag{14}$$

$$a_{v,l,c}^{\text{tran}} \Leftrightarrow \text{con}_=(\mathcal{T}_v^l, c) \tag{15}$$

As $\mathcal{A}_v^l$ variables appear only on the right side of implications in constraints (4) – (6) of the INVERSE encoding it is sufficient to connect their auxiliary by implications only. Whereas $\mathcal{T}_v^l$ variables appear on both sides of implications in (4) – (6); therefore they need to be connected by equivalences to their auxiliary variables.

Having above auxiliary variables, INVERSE encoding constraints can be easily expressed using them as follows:

- $a_{v,l,0}^{\text{tran}} \Rightarrow a_{v,v,l}^=$ (16)

    for every $v \in V$ and $l \in \{0,1,\dots,\eta-1\}$

- $a_{v,l,c}^{\text{tran}} \Rightarrow a_{u,l}^{\text{zero}} \wedge a_{u,v,l}^= \wedge a_{u,l,\sigma_u(v)+\deg_G(u)}^{\text{tran}}$ (17)

    for each $0 < c \leq \deg_G(v)$, $v \in V$ and $l \in \{0,1,\dots,\eta-1\}$, where $u = o_v^{-1}(c)$

- $a_{v,l,c}^{\text{tran}} \Rightarrow a_{u,l,\sigma_u(v)}^{\text{tran}}$ (18)

    for each $\deg_G(v) < c \leq 2 \cdot \deg_G(v)$, $v \in V$ and $l \in \{0,1,\dots,\eta-1\}$, where $u = \sigma_v^{-1}(\mathcal{T}_v^l - \deg_G(v))$

In the following space consumption of the INVERSE encoding only regular time layers are counted as asymptotically requirements of the initial and final time layers are dominated by the rest.

**Proposition 2** (*INVERSE ENCODING SIZE*). *The number of visible propositional variables in $F_{INV}(\eta, \Sigma)$ is $\mathcal{O}(\eta \cdot (|V| \cdot \lceil \log_2(\mu) \rceil + \sum_{v \in V} \lceil \log_2(\deg_G(v)) \rceil))$ and there are $\mathcal{O}(\eta \cdot (|V| + |E|))$ auxiliary variables; that is $\mathcal{O}(\eta \cdot (|V| \cdot \lceil \log_2(\mu) \rceil + \sum_{v \in V} \lceil \log_2(\deg_G(v)) \rceil + |E|))$ propositional variables in total. The number of clauses is $\mathcal{O}(\eta \cdot (|V| \cdot \lceil \log_2(\mu) \rceil + |E| \cdot \lceil \log_2(\mu) \rceil + \sum_{v \in V} \deg_G(v) \cdot (\lceil \log_2(\deg_G(v)) \rceil)))$.* ∎

**Proof.** To show the result we need just to calculate variables and clauses. The visible variables, that is, propositional variables representing $\mathcal{A}_v^l$ and $\mathcal{T}_v^l$ counts for $(\eta+1) \cdot |V| \cdot \lceil \log_2(\mu+1) \rceil$ and $\eta \cdot \sum_{v \in V} \lceil \log_2(2 \cdot \deg_G(v)+1) \rceil$ respectively. The number of auxiliary variables $a_{v,l}^{\text{zero}}$ is $(\eta+1) \cdot |V|$; the number of $a_{u,v,l}^=$ variables is $(\eta+1) \cdot |E|$; and the number of $a_{v,l,c}^{\text{tran}}$ variables is $2 \cdot \eta \cdot \sum_{v \in V} \deg_G(v)$ which is $4 \cdot \eta \cdot |E|$. Hence the total number of propositional variables is $(\eta+1) \cdot (|V| \cdot \lceil \log_2(\mu+1) \rceil + |V| + |E|) + \eta \cdot (\sum_{v \in V} \lceil \log_2(2 \cdot \deg_G(v)+1) \rceil + 4 \cdot |E|)$ which is $\mathcal{O}(\eta \cdot (|V| \cdot \lceil \log_2(\mu) \rceil + \sum_{v \in V} \lceil \log_2(\deg_G(v)) \rceil + |E|))$.

Let us calculate the number of clauses. A single constraint (13) develops into $\lceil\log_2(\mu+1)\rceil$ binary clauses; a single constraint (14) develops into $2 \cdot \lceil\log_2(\mu+1)\rceil$ ternary clauses; and a single constraint (15) develops into $\lceil\log_2(2 \cdot \deg_G(v)+1)\rceil$ binary clauses and one clause of arity $\lceil\log_2(2 \cdot \deg_G(v)+1)\rceil + 1$. There is as many as $\eta \cdot |V|$ constraints (13); $\eta \cdot |E|$ constraints (14); and $\eta \cdot \sum_{v \in V} \deg_G(v)$ constraints (15) which in total gives $\eta \cdot ((|V| + 2 \cdot |E|) \cdot \lceil\log_2(\mu+1)\rceil + \sum_{v \in V} \deg_G(v) \cdot (\lceil\log_2(2 \cdot \deg_G(v)+1)\rceil + 1))$ clauses (binary, ternary, and one multi-arity).

Constraints (16) count for $\eta \cdot |V|$ binary clauses, constraints (17) together with (18) count for $4 \cdot \eta \cdot \sum_{v \in V} \deg_G(v)$ binary clauses which is clearly dominated by the already calculated number of clauses. Hence, we have $\eta \cdot (|V| \cdot \lceil\log_2(\mu)\rceil + |E| \cdot \lceil\log_2(\mu)\rceil + \sum_{v \in V} \deg_G(v) \cdot (\lceil\log_2(\deg_G(v))\rceil))$ clauses. ∎

**Proposition 3** (*PATHS AND $F_{INV}(\eta, \Sigma)$ SATISFACTION*). *A set $\Pi = \{\pi_1, \pi_2, \dots, \pi_\mu\}$ of non-overlapping vertex disjoint paths in $\text{Exp}_T(G, \eta)$ so that $\pi_i$ connects $[\alpha_0(a_i), 0]$ with $[\alpha_+(a_i), \eta]$ for $i = 1, 2, \dots, \mu$ exists if and only if $F_{INV}(\eta, \Sigma)$ is satisfiable. Moreover, paths $\pi_1, \pi_2, \dots, \pi_\mu$ can be unambiguously constructed from satisfying valuation of $F_{INV}(\eta, \Sigma)$ and vice versa.* ∎

**Sketch of proof.** For simplicity, we will show the proposition over finite domain variables instead of bit-vectors. The equivalence between bit vectors and finite domain variables is can be seen directly from the translation of finite domain constraints to equivalent constraints over bit vectors.

Assume that there exists a collection of vertex disjoint paths $\Pi = \{\pi_1, \pi_2, \dots, \pi_\mu\}$, where $\pi_i$ connects $[\alpha_0(a_i), 0]$ with $[\alpha_+(a_i), \eta]$. Let $\pi_i = ([u_0, 0], [u_1, 1], [u_2, 2], \dots, [u_\eta, \eta])$, $u_l \in V$ for $l = 0, 1, \dots, \eta$ where $u_0 = \alpha_0(a_i)$ and $u_\eta = \alpha_+(a_i)$. We can set $\mathcal{A}_{u_0}^0 = i$, $\mathcal{A}_{u_1}^1 = i$, ..., $\mathcal{A}_{u_\eta}^\eta = i$. Transition variables are set according to traversed edges; that is, $\mathcal{T}_{u_0}^0 = \sigma_{u_0}(u_1)$, $\mathcal{T}_{u_1}^0 = \sigma_{u_1}(u_0) + \deg_G(u_1)$, $\mathcal{T}_{u_2}^1 = \sigma_{u_1}(u_2)$, $\mathcal{T}_{u_2}^1 = \sigma_{u_2}(u_1) + \deg_G(u_2)$, ..., $\mathcal{T}_{u_l}^l = \sigma_{u_l}(u_{l+1})$, $\mathcal{T}_{u_{l+1}}^l = \sigma_{u_{l+1}}(u_l) + \deg_G(u_{l+1})$, ..., $\mathcal{T}_{u_{\eta-1}}^{\eta-1} = \sigma_{u_{\eta-1}}(u_\eta)$, $\mathcal{T}_{u_\eta}^{\eta-1} = \sigma_{u_\eta}(u_{\eta-1}) + \deg_G(u_\eta)$. Other paths from $\Pi$ are processed in the same way. Observe that there is no conflict in setting the variables; that is, each variable is set at most once by the assignment; which is due to the fact that paths are vertex disjoin. Variables $\mathcal{A}_v^l$ and $\mathcal{T}_v^l$ that has not been set so far are set to 0. It is not difficult to check that constraints (4) – (6) as well as (7) – (11) are satisfied.

On the other hand, if there is a satisfying valuation of $F_{\eta-INV}(\Sigma)$ then we are able to reconstruct required vertex disjoint paths from it. Let $\pi_i = ([u_0, 0], [u_1, 1], [u_2, 2], \dots, [u_\eta, \eta])$ where $u_0 = \alpha_0(a_i)$, and $u_{l+1} = \sigma_{u_l}^{-1}(\mathcal{T}_{u_l}^l)$ for every $l = 0, 1, \dots, \eta - 1$ (it holds also that $u_l = \sigma_{u_{l+1}}^{-1}(\mathcal{T}_{u_{l+1}}^{l+1}) - \deg_G(u_{l+1})$). Transition state variables $\mathcal{T}_v^l$ that take just one value ensure that each vertex at each time layer needs to decide if it either is connected to a neighbor or accepts a connection from a neighbor (or is connected to itself). It is ensured that no intersection between selected paths appears as otherwise a vertex must have accepted connections from at least two sources or has to branch connections to at least two neighbors, which is both forbidden. A value of $\mathcal{A}_v^l$ variable is propagated to the next time layer only through the connection of the corresponding transition state variable $\mathcal{T}_v^l$. The fact that agents were propagated to their goals ensures that there must be a paths induced by transition state variables from initial positions of agents to their goal. ∎

The following theorem can be directly obtained by applying Proposition 1 and Proposition 3 which together justify solving of CPF via translation to SAT.

**Theorem 1** (SOLUTION OF $\Sigma$ AND $F_{INV}(\eta, \Sigma)$ SATISFACTION). *A solution of a CPF $\Sigma = (G, A, \alpha_0, \alpha_+)$ with $A = \{a_1, a_2, ..., a_\mu\}$ exists if and only if there exist $\eta \in \mathbb{N}$ for that formula $F_{\eta-INV}(\Sigma)$ is satisfiable.* ∎

### 4.2.2. ALL-DIFFERENT Propositional Encoding

Choosing location function instead of its inverse for representing arrangements of agents at individual time steps led to another encoding called ALL-DIFFERENT – the name comes from the fact that it is necessary to express the requirement that each vertex is occupied by at most one agent explicitly which is modeled by pair-wise differences between variables representing the arrangement. Again it is easier to express the encoding over finite domain state variables before it is transformed to propositional formula.

**Definition 6** (ALL-DIFFERENT ENCODING – $F_{DIFF}(\eta, \Sigma)$). Assume that a CPF $\Sigma = [G, A, \alpha_0, \alpha_+]$ with $G = (V, E)$ is given. An *ALL-DIFFERENT encoding* for CPF $\Sigma$ consists of finite domain variables $\mathcal{L}_a^l \in \{1, ..., n\}$ for every $a \in A$ and each time layer $l \in \{0, 1, ..., \eta\}$ to model locations of agents over time. Constraints are as follows:

- $\mathcal{L}_a^l = j \Rightarrow \mathcal{L}_a^{l+1} = j \vee \bigvee_{j' \in \{1,...,n\} | \{v_j, v_{j'}\} \in E} \mathcal{L}_a^{l+1} = j'$ (19)
  for every $a \in A$, $j \in \{1, 2, ..., n\}$ and $l \in \{0, 1, ..., \eta - 1\}$
  (agent $a$ moves along edges only or stay in a vertex)

- $\bigwedge_{b \in A | b \neq a} \mathcal{L}_a^{l+1} \neq \mathcal{L}_b^l$ for every $a \in A$ and $l \in \{0, 1, ..., \eta - 1\}$ (20)
  (target vertex of agent's $a$ move must be empty)

- AllDifferent($\mathcal{L}_{a_1}^l, \mathcal{L}_{a_2}^l, ..., \mathcal{L}_{a_\mu}^l$) for every $l \in \{0, 1, ..., \eta\}$ (21)
  (at most one agent reside in each vertex at each time step). □

Initial and goal arrangements will be expressed though the following constraints:

- $\mathcal{L}_a^0 = j$  for $a \in A$ with $\alpha_0(a) = v_j$  } Initial locations (22)
- $\mathcal{L}_a^\eta = j$  for $a \in A$ with $\alpha_+(a) = v_j$  } Goal locations (23)

Again, finite domain state variables $\mathcal{L}_a^l$ are represented as a bit vector (vector of propositional variables) using binary encoding. That is, $\lceil \log_2 |V| \rceil$ propositional variables are introduced for each $\mathcal{L}_a^l$ variable. The resulting formula in CNF will be denoted as $F_{DIFF}(\eta, \Sigma)$.

AllDifferent($\mathcal{L}_{a_1}^l, \mathcal{L}_{a_2}^l, ..., \mathcal{L}_{a_\mu}^l$) constraint requires that all the involved variables are assigned different values; that is, $\bigwedge_{j,k \in \{1,2,...,\mu\} | j < k} \mathcal{L}_{a_j}^l \neq \mathcal{L}_{a_k}^l$. Differences between finite domain state variables are encoded using the scheme introduced in [1]. The scheme is used to encode constraints (20) as well as (21). Inequality between variables $\mathcal{L}_{a_j}^l$ and $\mathcal{L}_{a_k}^l$ is expressed in the scheme by the following clauses. Auxiliary variables $d_{j,k}^l$ representing difference at individual bits are introduced.

$$\text{var}_{\neq}\left(\mathcal{L}_{a_j}^l, \mathcal{L}_{a_k}^l\right) = \bigvee_{\hat{\imath}=0}^{\lceil \log_2 n \rceil - 1} d_{j,k}^l \tag{24}$$

where $\bigwedge_{\hat{\imath}=0}^{\lceil \log_2 n \rceil - 1} \left(\neg d_{j,k}^l \vee \neg \mathcal{L}_{a_j}^l[\hat{\imath}] \vee \mathcal{L}_{a_k}^l[\hat{\imath}]\right) \wedge \left(\neg d_{j,k}^l \vee \mathcal{L}_{a_j}^l[\hat{\imath}] \vee \neg \mathcal{L}_{a_k}^l[\hat{\imath}]\right)$

Conditional equality disjunction (19) is encoded by introducing auxiliary propositional variables to represent equalities between bit vectors. For each $j \in \{1,2,\ldots,n\}$ (that is, for each vertex), agent $a \in A$, and time layer $l \in \{0,1,\ldots,\eta\}$, an auxiliary variable $e_{a,j}^l$ which stands for equality $\mathcal{L}_a^l = j$ is introduced. The link between auxiliary variables $e_{a,j}^l$ and actual equalities is established through the following constraint:

$$e_{a,j}^l \Leftrightarrow \text{con}_=(\mathcal{L}_a^l, j) \tag{25}$$

Then moving along edges – constraints (19) – can be easily expressed as single clause over auxiliary variables:

$$e_{a,j}^l \Rightarrow \neg e_{a,j}^{l+1} \vee \bigvee_{j \in \{1,\ldots,n\} | \{v_j, v_{\hat{\jmath}}\} \in E} e_{a,\hat{\jmath}}^{l+1} \tag{26}$$

Again, the space consumption of the ALL-DIFFERENT encoding will be calculated for regular time layers only.

**Proposition 4** (ALL-DIFFERENT ENCODING SIZE). *The number of visible propositional variables in $F_{DIFF}(\eta, \Sigma)$ is $\mathcal{O}(\eta \cdot \mu \cdot \lceil \log_2 |V| \rceil)$ and there are $\mathcal{O}(\eta \cdot \mu \cdot |V|)$ auxiliary variables; that is, the number of variables is $\mathcal{O}(\eta \cdot \mu \cdot |V|)$. The number of clauses is $\mathcal{O}\left(\eta \cdot \lceil \log_2 |V| \rceil \cdot \left(\binom{\mu}{2} + \mu \cdot |V|\right)\right)$.* ∎

**Proof.** Let us calculate the number of variables and clauses. Each variable $\mathcal{L}_a^l$ is represented by $\log_2 |V|$ variables and the number of $\mathcal{L}_a^l$ variables is $(\eta + 1) \cdot \mu$. For each $\mathcal{L}_a^l$ variable and its value, an auxiliary variable is introduced. As $\mathcal{L}_a^l$ can take $|V|$ values, we get the result that there are $(\eta + 1) \cdot \mu \cdot (\log_2 |V| + |V|)$ variables which is $\mathcal{O}(\eta \cdot \mu \cdot |V|)$.

A single time layer requires as many as $\binom{\mu}{2}$ inequalities between all pairs of $\mathcal{L}_a^l$ variables corresponding to distinct agents to model the AllDifferent constraint from (21). Each inequality is modeled by $2 \cdot \lceil \log_2 |V| \rceil$ ternary clauses plus one clause of arity $\lceil \log_2 |V| \rceil$. This is in total $(\eta + 1) \cdot \binom{\mu}{2} \cdot (2 \cdot \lceil \log_2 |V| \rceil + 1)$ clauses.

Next, we need as many as $\binom{\mu}{2}$ inequalities between $\mathcal{L}_a^l$ variables from two consecutive time layers (constraint (20)) which adds the same number of $(\eta + 1) \cdot \binom{\mu}{2} \cdot (2 \cdot \lceil \log_2|V| \rceil + 1)$ clauses again.

Links between auxiliary variables $e_{a,j}^l$ and actual equalities (25) they represent need $\lceil \log_2|V| \rceil$ binary clauses plus one clause of arity $\lceil \log_2|V| \rceil + 1$, which is $(\eta + 1) \cdot \mu \cdot |V| \cdot (\lceil \log_2|V| \rceil + 1)$ in total.

Finally, constraints expressing that agents move along edges only (26) contribute to each vertex $v_j$ in $\text{Exp}_T(G, \eta)$ at given time layer except the last one by $\mu$ clauses of arity $\deg_G(v_j) + 2$ which is $\eta \cdot \mu \cdot |V|$ clauses in total.

Altogether we have $(\eta + 1) \cdot \left( \binom{\mu}{2} \cdot (2 \cdot \lceil \log_2|V| \rceil + 1) + \mu \cdot |V| \cdot (\lceil \log_2|V| \rceil + 1) \right) + \eta \cdot \mu \cdot |V|$ clauses in $F_{DIFF}(\eta, \Sigma)$ encoding which is $\mathcal{O}\left( \eta \cdot \lceil \log_2|V| \rceil \cdot \left( \binom{\mu}{2} + \mu \cdot |V| \right) \right)$. ∎

**Proposition 5** (*PATHS AND $F_{DIFF}(\eta, \Sigma)$ SATISFACTION*). *A set $\Pi = \{\pi_1, \pi_2, \ldots, \pi_\mu\}$ of non-overlapping vertex disjoint paths in $\text{Exp}_T(G, \eta)$ so that $\pi_i$ connects $[\alpha_0(a_i), 0]$ with $[\alpha_+(a_i), \eta]$ for $i = 1, 2, \ldots, \mu$ exists if and only if $F_{DIFF}(\eta, \Sigma)$ is satisfiable. Moreover, paths $\pi_1, \pi_2, \ldots, \pi_\mu$ can be unambiguously constructed from satisfying valuation of $F_{DIFF}(\eta, \Sigma)$ and vice versa.* ∎

**Sketch of proof.** For simplicity, we will work on the level of finite domain state variables. Assume that non-overlapping vertex disjoint paths $\pi_1, \pi_2, \ldots, \pi_\mu$ exist in $\text{Exp}_T(G, \eta)$. The satisfying valuation of $F_{DIFF}(\eta, \Sigma)$ can be directly constructed from these paths. Let $\pi_i = ([u_0, 0], [u_1, 1], [u_2, 2], \ldots, [u_\eta, \eta]), u_l \in V$ for $l = 0, 1, \ldots, \eta$ where $u_0 = \alpha_0(a_i)$ and $u_\eta = \alpha_+(a_i)$. Then finite domain state variables will be set as follows: $\mathcal{L}_{a_i}^0 = u_0$, $\mathcal{L}_{a_i}^1 = u_1$, ..., $\mathcal{L}_{a_i}^\eta = u_\eta$ for every $i = 1, 2, \ldots, \mu$. The assumptions that paths were vertex disjoint and non-overlapping ensure that constraints (21) and (20) respectively are satisfied. Consecutive vertices in paths are connected by directed edges that correspond to edges in $G$. Hence, constraints (19) are satisfied.

Assume on the other hand that we have a satisfying valuation of $F_{DIFF}(\eta, \Sigma)$. We can immediately set $\pi_i = ([\mathcal{L}_{a_i}^0, 0], [\mathcal{L}_{a_i}^1, 1], [\mathcal{L}_{a_i}^2, 2], \ldots, [\mathcal{L}_{a_i}^\eta, \eta])$ for every $i = 1, 2, \ldots, \mu$. Satisfaction of constraints (19) ensures that constructed sequences of vertices are paths in $\text{Exp}_T(G, \eta)$ which are moreover vertex disjoint and non-overlapping due to constraints (21) and (20). ∎

CPF solving via $F_{DIFF}(\eta, \Sigma)$ satisfaction is justified by the following theorem which can be shown by combining Proposition 1 and just proven Proposition 5 (from $F_{DIFF}(\eta, \Sigma)$ satisfaction non-overlapping vertex disjoint paths can be obtained which correspond to CPF solution).

**Theorem 2** (*SOLUTION OF $\Sigma$ AND $F_{DIFF}(\eta, \Sigma)$ SATISFACTION*). *A solution of a CPF $\Sigma = (G, A, \alpha_0, \alpha_+)$ exists if and only if there exist $\eta \in \mathbb{N}$ for that formula $F_{DIFF}(\eta, \Sigma)$ is satisfiable.* ∎

### 4.2.3. MATCHING Propositional Encoding

We observed that vertex disjoint non-overlapping paths in time expansion graph resemble a *commodity flow* [1] in a network of time expansion graph where vertices and edges are assigned unit capacities. The intuition is that edges included into paths should be saturated by one unit of the flow. Such setting conveys commodity from each initial vertex to each goal vertex. However, the correspondence between paths of required properties and flow works in one direction only. The flow reflects well the requirement that paths should be vertex disjoint but does not simulate non-overlapping between paths as well as the correct interconnection between initial and goal vertex of the same agents (the flow may interconnect initial and goal vertices of two distinct agents).

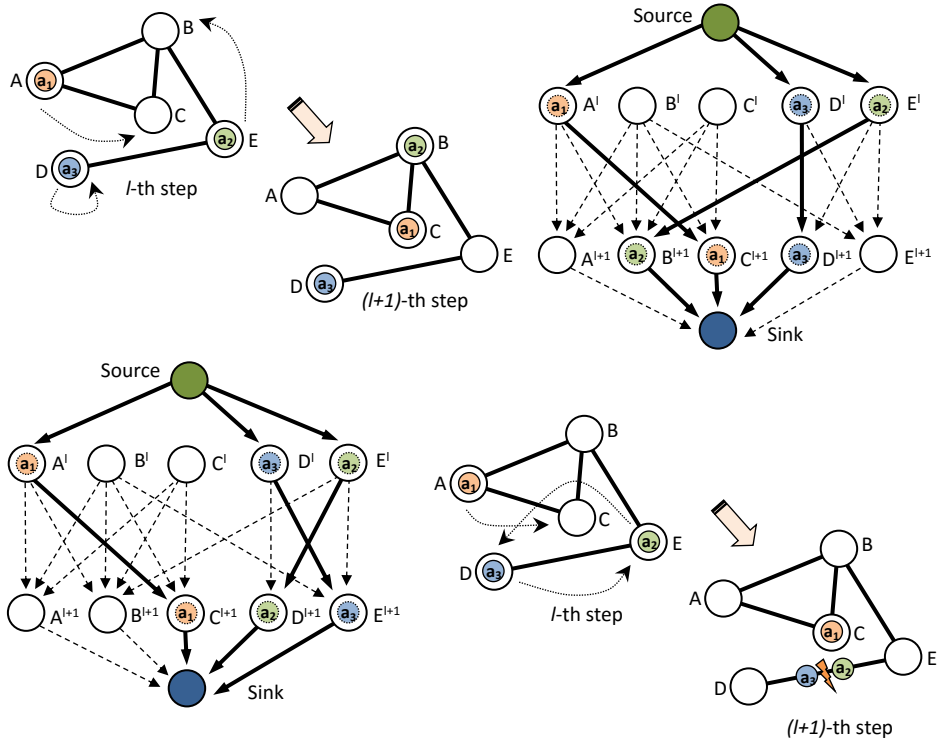

**Figure 4.** *Correspondence of agent movement and flow in bipartite graph.* Movement between time steps $l$ and $l + 1$ and corresponding flow in bipartite graph made of $l$-th and $(l + 1)$-th time layer where vertices and edges are assigned unit capacities is shown. Valid movement always induces flow in which saturated edges are non-overlapping; that is, $\{A, E\} \cap \{B, C\} = \emptyset$ (*upper part*). On the other hand, flow does not necessarily induce non-overlapping edges which may result in invalid movement (*lower part*).

The design of the MATCHING encoding will follow the intuition suggested by single commodity flows. It will be divided into two parts – the first part, called FLOW part, will check the existence of a flow that generates non-overlapping vertex disjoint paths. This part can be regarded as an encoding of a relaxed CPF with anonymous agents where we care about relocation of a group of agents to a set of goal vertices but we don't care about what particular agent arrives at particular goal vertex (generated paths may interconnect initial and goal vertices of distinct agents). The second part, called MAPPING part, of the encoding maps distinguishable agents to paths marked out by the flow, which eventually override the relaxation from the first part of the encoding. The encoding should allow fast testing of the existence of non-overlapping flow to enable using this test as a heuristic since its existence is necessary condition for existence of a solution.

**Definition 7** (MATCHING ENCODING – $F_{\eta-MATCH}^{FLOW}(\Sigma)$). A *FLOW part of the MATCHING encoding* of given CPF $\Sigma = [G, A, \alpha_0, \alpha_+]$ with $G = (V, E)$ consists of a propositional variable for each vertex and edge in the time expansion graph that model its saturation by the flow. That is, propositional variable $\mathcal{M}_v^l$ is introduced for every $l = 0, 1, \ldots, \eta$ and $v \in V$ and propositional variables $\mathcal{E}_{u,v}^l$ and $\mathcal{E}_u^l$ are introduced for every $l = 0, 1, \ldots, \eta$ and $\{u, v\} \in E$ and $u \in V$ respectively. Constraints enforce that variables set to *TRUE* form a non-overlapping flow:

- $\mathcal{E}_{u,v}^l \Rightarrow \mathcal{M}_u^l \wedge \mathcal{M}_v^{l+1}$     for every $\{u, v\} \in E$     (25)
  and $l \in \{0, 1, \ldots, \eta - 1\}$,
  $\mathcal{E}_u^l \Rightarrow \mathcal{M}_u^l \wedge \mathcal{M}_u^{l+1}$     for every $u \in V$ and $l \in \{0, 1, \ldots, \eta - 1\}$
  (if an edge is selected into flow then its endpoints are selected as well)

- $\mathcal{E}_u^l + \sum_{v | \{u,v\} \in E} \mathcal{E}_{u,v}^l \leq 1$     for every $u \in V$ and $l \in \{0, 1, \ldots, \eta - 1\}$,     (26)
  $\mathcal{E}_v^l + \sum_{u | \{u,v\} \in E} \mathcal{E}_{u,v}^l \leq 1$     for every $v \in V$ and $l \in \{0, 1, \ldots, \eta - 1\}$,
  (at most one incoming and outgoing edge is selected into flow)

- $\mathcal{M}_u^l \Rightarrow \mathcal{E}_u^l \vee \bigvee_{v | \{u,v\} \in E} \mathcal{E}_{u,v}^l$     for every $u \in V$ and $l \in \{0, 1, \ldots, \eta - 1\}$,     (27)
  $\mathcal{M}_v^{l+1} \Rightarrow \mathcal{E}_v^l \vee \bigvee_{u | \{u,v\} \in E} \mathcal{E}_{u,v}^l$     for every $v \in V$ and $l \in \{0, 1, \ldots, \eta - 1\}$,
  (if a vertex is selected into flow then at least one outgoing and
  incoming edge must be selected as well)

- $\mathcal{E}_{u,v}^l \Rightarrow \neg \mathcal{M}_v^l$     for every $\{u, v\} \in E$     (28)
  and $l \in \{0, 1, \ldots, \eta - 1\}$,
  (source and target vertices of non-trivial moves must be disjoint). □

The second part of the encoding where individual distinguishable agents manifest themselves is introduced in the following definition.

**Definition 8** (MATCHING ENCODING – $F_{\eta-MATCH}^{MAP}(\Sigma)$). A *MAPPING part of the MATCHING encoding* of given CPF $\Sigma = [G, A, \alpha_0, \alpha_+]$ with $G = (V, E)$ consists of a finite domain variable $\mathcal{A}_v^l \in \{0, 1, \ldots, \mu\}$ for each vertex $v \in V$ and every time layer $l = 0, 1, \ldots, \eta$ to model agent occurrence in a vertex. Constraints interconnect the MAPPING part with FLOW part so that actual agents follow paths indicated by the flow:

- $\mathcal{E}_{u,v}^l \Rightarrow \mathcal{A}_u^l = \mathcal{A}_v^{l+1}$     for every $\{u, v\} \in E$     (29)
  and $l \in \{0, 1, \ldots, \eta - 1\}$,

(if an edge is saturated by the flow then the same agent appears at its both ends)

- $\mathcal{A}_u^l \neq 0 \Rightarrow \mathcal{M}_u^l$        for every $u \in V$        (30)
and $l \in \{0,1,\ldots,\eta\}$

(if an agent occurs in a vertex then the vertex is saturated by the flow) □

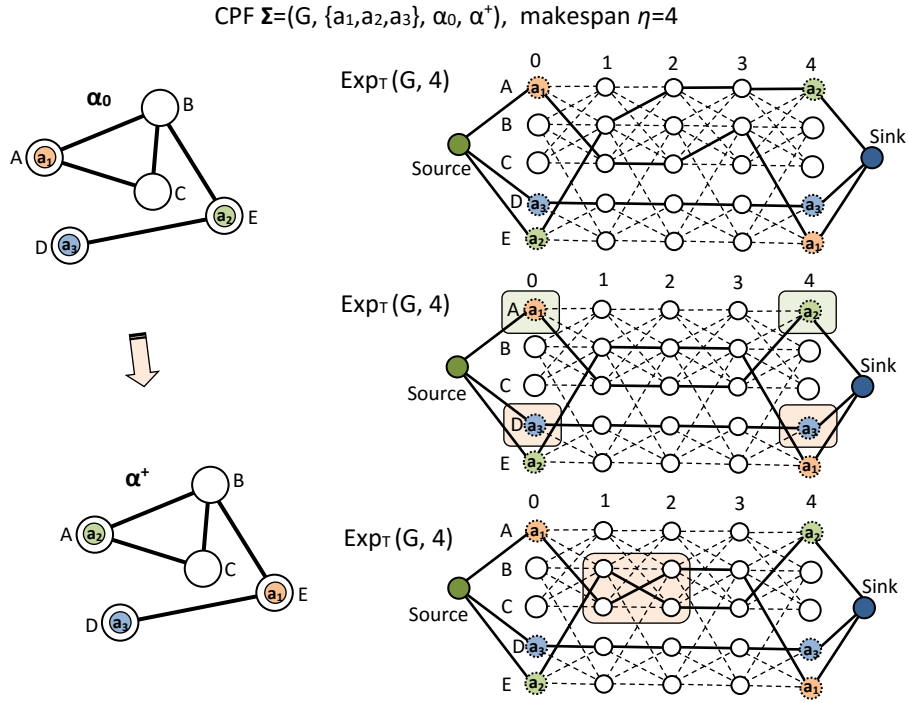

**Figure 5.** *Non-overlapping vertex disjoint paths in time expansion graph depicted and single commodity flow correspondence.* Edges and vertices in time expansion graph are assumed to have unit capacities. A correct flow can be reconstructed from vertex disjoint non-overlapping paths (*upper right part*). On the other hand, flow does not necessarily correspond to paths of required properties (*middle right part* shows connection of initial and goal vertices of different agents and lower right part shows overlapping paths between time layers 1 and 2).

As in the previous encodings $\mathcal{A}_v^l$ variables having $\mu + 1$ states are represented by $\lceil \log_2(\mu + 1) \rceil$ propositional variables using binary encoding. Initial and goal arrangements will be expressed though the following constraints:

- $\mathcal{A}_u^0 = i \wedge \mathcal{M}_u^0$     for $u \in V$ if there is $i \in \{1,2,\ldots,\mu\}$     (31)
such that $\alpha_0(a_i) = u$     Initial locations
- $\mathcal{A}_u^0 = 0 \wedge \neg \mathcal{M}_u^0$     for $u \in V$ if $(\forall a \in A)\alpha_0(a) \neq u$     (32)

- $\mathcal{A}_u^\eta = i \wedge \mathcal{M}_u^\eta$     for $u \in V$ if there is $i \in \{1,2,...,\mu\}$ such that $\alpha_+(a_i) = u$    } Goal locations    (33)
- $\mathcal{A}_u^\eta = 0 \wedge \neg \mathcal{M}_u^\eta$    for $u \in V$ if $(\forall a \in A)\alpha_+(a) \neq u$    (34)

The resulting formula of the MATCHING encoding in CNF is a conjunction of the FLOW part, MAPPING part, and boundary conditions and is denoted as $F_{\eta-MATCH}(\Sigma)$. To obtain CNF it is necessary to rewrite (26) as clauses. That is, for example $\mathcal{E}_u^l + \sum_{v|\{u,v\}\in E} \mathcal{E}_{u,v}^l \leq 1$ is rewritten as a conjunction of clauses that forbid all pairs of involved variables to be set to *TRUE* simultaneously:

$$\bigwedge_{v,w|\{u,v\}\in E \wedge \{u,w\}\in E \wedge v \neq w} \neg \mathcal{E}_{u,v}^l \vee \neg \mathcal{E}_{u,w}^l$$
$$\bigwedge_{v|\{u,v\}\in E} \neg \mathcal{E}_{u,v}^l \vee \neg \mathcal{E}_u^l \quad (35)$$

Binary encoded variables $\mathcal{A}_v^l$ are not involved in any complex relation – only conditional equality between these variables are introduced while all other modeling issues concerning validity conditions are done in the FLOW part of the encoding.

The conditional equality between $\mathcal{A}_u^l$ and $\mathcal{A}_v^{l+1}$ (25) can be expressed using construct introduced earlier:

$$\mathcal{E}_{u,v}^l \Rightarrow \text{var}_=(\mathcal{A}_u^l, \mathcal{A}_v^{l+1}) \quad (36)$$

Constraint (26) can be also easily rewritten as follows:

$$\bigwedge_{\hat{\imath}=0}^{\lceil \log_2(\mu+1) \rceil - 1} \neg \mathcal{A}_u^l[\hat{\imath}] \vee \mathcal{M}_u^l \quad (37)$$

**Proposition 6** (*MATCHING ENCODING SIZE*). *The number of propositional variables in $F_{\eta-MATCH}(\Sigma)$ is $\mathcal{O}(\eta \cdot (|E| + |V| \cdot \lceil \log_2(\mu) \rceil))$. The number of clauses is $\mathcal{O}\left(\eta \cdot \left((|V| + |E|) \cdot \lceil \log_2(\mu) \rceil + \sum_{v \in V} \binom{\deg_G(v)}{2}\right)\right)$.* ∎

**Proof.** The FLOW part of the MATCHING encoding has a propositional variable $\mathcal{M}_v^l$ for each vertex $v \in V$ and time layer $l \in \{0,1,...,\eta\}$ and $\mathcal{E}_{u,v}^l$ for each edge $\{u,v\} \in E$ and time layer, which in total makes $(\eta + 1) \cdot (|V| + |E|)$ propositional variables. Further, we have a vector of $\lceil \log_2(\mu + 1) \rceil$ propositional variables representing $\mathcal{A}_v^l$ for each vertex and time layer in the MAPPING part. This in total makes another $(\eta + 1) \cdot |V| \cdot \lceil \log_2(\mu + 1) \rceil$ variables. Altogether, there are $(\eta + 1) \cdot (|E| + |V| + |V| \cdot \lceil \log_2(\mu + 1) \rceil)$ variables which is $\mathcal{O}(\eta \cdot (|E| + |V| \cdot \lceil \log_2(\mu) \rceil))$.

Constraints (25) develops into $\eta \cdot (|V| + |E|)$ ternary clauses. Constraints (26) develop into $2 \cdot \eta \cdot \sum_{v \in V} \binom{\deg_G(v)+1}{2}$ binary clauses as indicated by (31). Constraints (27) introduce two clauses of length $\deg_G(v) + 1$ for each vertex and time layer; that is, $2 \cdot \eta \cdot |V|$ clauses are added. Finally, constraints (28) add a binary clause for each vertex and time layer, which is again dominated by previous expressions. Conditional equality between two bit vectors in (29) develops into $2 \cdot \lceil \log_2(\mu + 1) \rceil$ ternary clauses while the equality is introduced for each edge and time layer; that is, $2 \cdot \eta \cdot |E| \cdot \lceil \log_2(\mu + 1) \rceil$ ternary clauses are added. It is easy to observe that expression (30) represents $(\eta + 1) \cdot |E| \cdot \lceil \log_2(\mu + 1) \rceil$ binary clauses. Altogether, there are $\eta \cdot (3 \cdot |V| + |E| + 2 \cdot \sum_{v \in V} \binom{\deg_G(v)+1}{2}) + 2 \cdot |E| \cdot \lceil \log_2(\mu + 1) \rceil) + (\eta + 1) \cdot |E| \cdot \lceil \log_2(\mu + 1) \rceil$ clauses which is $\mathcal{O}\left(\eta \cdot \left((|V| + |E|) \cdot \lceil \log_2(\mu) \rceil + \sum_{v \in V} \binom{\deg_G(v)}{2}\right)\right)$. ∎

**Proposition 7** (*PATHS AND $F_{MATCH}(\eta, \Sigma)$ SATISFACTION*). *A set $\Pi = \{\pi_1, \pi_2, \ldots, \pi_\mu\}$ of non-overlapping vertex disjoint paths in $\mathrm{Exp}_T(G, \eta)$ so that $\pi_i$ connects $[\alpha_0(a_i), 0]$ with $[\alpha_+(a_i), \eta]$ for $i = 1, 2, \ldots, \mu$ exists if and only if $F_{MATCH}(\eta, \Sigma)$ is satisfiable. Moreover, paths $\pi_1, \pi_2, \ldots, \pi_\mu$ can be unambiguously constructed from satisfying valuation of $F_{MATCH}(\eta, \Sigma)$ and vice versa.* ∎

**Sketch of proof.** We will work at the level of finite domain state variables $\mathcal{A}_v^l$ instead of bit vectors to simplify the proof.

Assume that non-overlapping vertex disjoint paths $\pi_1, \pi_2, \ldots, \pi_\mu$ exist so that $\pi_i$ connects $[\alpha_0(a_i), 0]$ with $[\alpha_+(a_i), \eta]$ in $\mathrm{Exp}_T(G, \eta)$. Let $\pi_i = ([u_0, 0], [u_1, 1], [u_2, 2], \ldots, [u_\eta, \eta])$, with $u_l \in V$ for $l = 0, 1, \ldots, \eta$ where $u_0 = \alpha_0(a_i)$ and $u_\eta = \alpha_+(a_i)$. The satisfying valuation of $F_{MATCH}(\eta, \Sigma)$ can be easily constructed by setting $\mathcal{A}_{u_0}^0 = i$, $\mathcal{A}_{u_1}^1 = i$, ..., $\mathcal{A}_{u_\eta}^\eta = i$. Next, variables representing flow $\mathcal{M}_{u_0}^0, \mathcal{M}_{u_1}^1, \ldots, \mathcal{M}_{u_\eta}^\eta$ are set to $TRUE$ and $\mathcal{E}_{u_0, u_1}^0$, $\mathcal{E}_{u_1, u_2}^1$, ..., $\mathcal{E}_{u_{\eta-1}, u_\eta}^{\eta-1}$ are set to $TRUE$ as well (the convention that $\mathcal{E}_{u_l, u_{l+1}}^l \equiv \mathcal{E}_{u_l}^l$ if $u_l = u_{l+1}$ is used here). Now, observe that all the constraints are satisfied. The interconnection between the FLOW and the MAPPING part (constraints (25) and (26)) is satisfied by the construction so we just need to check constraints in the FLOW part of the encoding. Propagation of the flow from edges to vertices (constraints (24)) is also ensured by the construction. The fact that original paths are vertex disjoint ensures validity of constraints (25) and (26) which together enforce selection of exactly one incoming and one outgoing edge through setting $\mathcal{E}_{u,v}^l$ variables for each vertex saturated by the flow indicated by $\mathcal{M}_v^l$ variable set to $TRUE$. Finally, the non-overlapping property of paths is directly translated to satisfaction of constraints (28). Initial and goal location constraints are trivially satisfied. Altogether, $F_{MATCH}(\eta, \Sigma)$ is satisfied by constructed valuation of its variables.

Now let us check the opposite implication. Assume that $F_{MATCH}(\eta, \Sigma)$ is satisfiable. Let $\pi_i = ([u_0, 0], [u_1, 1], [u_2, 2], \ldots, [u_\eta, \eta])$ such that $u_0 = \alpha_0(a_i)$ and $\mathcal{E}_{u_l, u_{l+1}}^l$ is $TRUE$ for each $l = 0, 1, \ldots, \eta - 1$. This can be done due to constraints (25) - (27) that propagate flow from the initial positions in the first time layer towards final layer. We shall verify that paths constructed in this way have required properties – are vertex disjoint non-overlapping and interconnects initial and goal positions of agents. FLOW part of the encoding ensures that constructed paths are vertex disjoint and non-overlapping. We need just to add non-overlapping to already checked flow propagation. The non-overlapping is established by constraints (28). However, the FLOW part does not ensure that $u_\eta = \alpha_+(a_i)$; satisfaction of the FLOW part alone may result in a path that interconnects initial and goal positions of two distinct agents. This is corrected by constraints included in the MAPPING part of the encoding. These constraints propagate agent $a_i$ along edges $\{u_l, u_{l+1}\}$ and eventually force it to appear in $u_\eta$ where goal constraints (33) and (34) ensure that agent $a_i$ arrives to the right vertex. ■

By combining just proven Proposition 7 and correspondence between non-overlapping vertex disjoint paths in $\text{Exp}_T(G, \eta)$ the following theorem can be immediately obtained.

**Theorem 3** (SOLUTION OF $\Sigma$ AND $F_{MATCH}(\eta, \Sigma)$ SATISFACTION). *A solution of a CPF $\Sigma = (G, A, \alpha_0, \alpha_+)$ exists if and only if there exist $\eta \in \mathbb{N}$ for that formula $F_{MATCH}(\eta, \Sigma)$ is satisfiable.* ■

### 4.2.4. DIRECT/SIMPLIFIED Propositional Encoding

As all the previous encodings of CPF used binary representation of agent occurrence in a vertex in some form, we also considered encoding of CPF that expresses agent occurrences in vertices directly. That is, there will be single propositional variable that encodes occurrence of a given agent in a given vertex at a given time-step. The resulting CPF encoding will be called DIRECT. While the design of variables is extremely simple in the DIRECT encoding, the set of constraints is more complex as summarized in the following definition [30].

**Definition 9** (DIRECT ENCODING – $F_{DIR}(\eta, \Sigma)$). Assume that a CPF $\Sigma = [G, A, \alpha_0, \alpha_+]$ with $G = (V, E)$ is given. A *DIRECT encoding* for CPF $\Sigma$ consists of propositional variables $\mathcal{X}_{a,v}^l$ for every $a \in A$, $v \in V$, and time layer $l \in \{0, 1, \ldots, \eta\}$ to model occurrences of agents in vertices over time. The interpretation is that $\mathcal{X}_{a,v}^l$ is assigned *TRUE* if and agent $a$ appears in vertex $v$ at time step $l$. The following constrains ensure satisfaction of validity conditions between consecutive arrangements of agents:

- $\bigwedge_{u,v \in V, u \neq v} \neg \mathcal{X}_{a,u}^l \vee \neg \mathcal{X}_{a,v}^l$     for every $l \in \{0,1,\ldots,\eta\}$     (38)
  $\bigvee_{v \in V} \mathcal{X}_{a,v}^l$     and $a \in A$
  (an agent is placed in exactly one vertex at each time step)

- $\bigwedge_{a,b \in A, a \neq b} \neg \mathcal{X}_{a,v}^l \vee \neg \mathcal{X}_{b,v}^l$     for every $l \in \{0,1,\ldots,\eta\}$     (39)
  and $v \in V$
  (at most one agent is placed in each vertex at each time step)

- $\mathcal{X}_{a,v}^l \Rightarrow \mathcal{X}_{a,v}^{l+1} \vee \bigvee_{u \in V, \{v,u\} \in E} \mathcal{X}_{a,u}^{l+1}$     for every $l \in \{0,1,\ldots,\eta-1\}$,     (40)
  $\mathcal{X}_{a,v}^{l+1} \Rightarrow \mathcal{X}_{a,v}^l \vee \bigvee_{u \in V, \{v,u\} \in E} \mathcal{X}_{a,u}^l$     $v \in V$, and $a \in A$
  (an agent relocates to some of its neighbors or makes no move)

- $\mathcal{X}_{a,v}^l \wedge \mathcal{X}_{a,u}^{l+1} \Rightarrow \bigwedge_{b \in A} \neg \mathcal{X}_{b,u}^l \wedge \bigwedge_{b \in A} \neg \mathcal{X}_{b,v}^{l+1}$     (41)
           for every $l \in \{0,1,\ldots,\eta-1\}$, $u, v \in V$
           such that $\{u, v\} \in E$ and $a \in A$
  (target vertex of a move must be vacant and the source vertex will be vacant after the move is performed). □

Initial and goal arrangements will be expressed though the following constraints:

- $\mathcal{X}_{a,v}^0$           for $v \in V$ if there is $a \in A$                                     (42)
               such that $\alpha_0(a) = v$    Initial locations
- $\neg \mathcal{X}_{a,v}^0$      otherwise                                                            (43)
- $\mathcal{X}_{a,v}^\eta$         for $v \in V$ if there is $a \in A$
              such that $\alpha_+(a) = v$    Goal locations                      (44)
- $\neg \mathcal{X}_{a,v}^\eta$      otherwise                                                            (45)

The resulting DIRECT encoding formula in CNF will be denoted as $F_{DIR}(\eta, \Sigma)$. It can be easily observed that the vacancy of target vertex and source vertex before and after the move (constraint (41)) is quite repetitive as the right side of the implication is independent of agent $a$. Therefore, the encoding is enhanced by introducing auxiliary variables $\mathcal{E}_u^l$ for each vertex $u \in V$ and time layer $l \in \{0,1,\dots,\eta\}$ that represent vacancy of vertex $u$ at time step $l$. Semantics of $\mathcal{E}_u^l$ variables is represented by the following constraint:

- $\mathcal{E}_u^l \Rightarrow \bigwedge_{a \in A} \neg \mathcal{X}_{a,u}^l$          for every $l \in \{0,1,\dots,\eta\}$ and $u \in V$        (46)
  (in an empty vertex no agent can appear at given time)

The repetitive part in constraint (41) can be then replaced by its version with auxiliary variables as follows:

- $\mathcal{X}_{a,v}^l \wedge \mathcal{X}_{a,u}^{l+1} \Rightarrow \mathcal{E}_u^l \wedge \mathcal{E}_v^{l+1}$       for every $l \in \{0,1,\dots,\eta-1\}$, $u, v \in V$    (47)
                                                   such that $\{u,v\} \in E$ and $a \in A$.

The resulting encoding with auxiliary variables will be called SIMPLIFIED and the corresponding CNF formula will be denoted as $F_{SIM}(\eta, \Sigma)$.

**Proposition 8** (*DIRECT/SIMPLIFIED ENCODING SIZE*). *The number of propositional variables in $F_{DIR}(\eta, \Sigma)$ is $\mathcal{O}(\eta \cdot \mu \cdot |V|)$. The number of clauses is $\mathcal{O}(\eta \cdot (\mu \cdot |V|^2 + \mu^2 \cdot |V| + \mu^2 \cdot |E|))$. $F_{SIM}(\eta, \Sigma)$ contains additional $\mathcal{O}(\eta \cdot |V|)$ propositional variables while the total number of clauses is $\mathcal{O}(\eta \cdot (\mu \cdot |V|^2 + \mu^2 \cdot |V| + \mu \cdot |E|))$.* ∎

**Proof.** It is easy to see that there are exactly $(\eta + 1) \cdot \mu \cdot |V|$ variables $\mathcal{X}_{a,v}^l$ and $\eta \cdot |V|$ $\mathcal{E}_u^l$ variables just by calculating their index scopes which gives us the result regarding the number of propositional variables in $F_{DIR}(\eta, \Sigma)$ and $F_{SIM}(\eta, \Sigma)$.

Every time layer and agent adds $\binom{|V|}{2}$ binary clauses and one $|V|$-ary clause within constraints (38). Thus, we have $(\eta + 1) \cdot \mu \cdot \binom{|V|}{2}$ binary clauses and $(\eta + 1) \cdot \mu$ $|V|$-ary from writing constraints (38) as clauses in total. The similar calculation can be done for constraints (39); we have $\binom{\mu}{2}$ binary clauses for each time layer and a vertex; that is, $(\eta + 1) \cdot |V| \cdot \binom{\mu}{2}$ binary clauses in total.

There are two $(\deg_G(v) + 2)$-ary clauses for every vertex $v$ in every time layer except the last one and for every agent from constraints (40), which in total gives $\eta \cdot \mu \, (\deg_G(v) + 2)$-ary clauses for each vertex $v \in V$. That is, $\eta \cdot \mu \cdot |V|$ clauses in total.

Note that each implication in constraint (41) develops into $2 \cdot \mu$ ternary clauses. There are $|E|$ such groups of clauses for every agent and a time layer except the last one. Thus, $2 \cdot \eta \cdot \mu^2 \cdot |E|$ ternary clauses are needed in total for expressing constraints (41).

Altogether, the total number of clauses in $F_{DIR}(\eta, \Sigma)$ is $(\eta + 1) \cdot \left(\mu \cdot \left(\binom{|V|}{2} + 1\right) + |V| \cdot \binom{\mu}{2}\right) + 2 \cdot \eta \cdot \mu \cdot (|V| + \mu \cdot |E|)$ which is $\mathcal{O}\left(\eta \cdot (\mu \cdot |V|^2 + \mu^2 \cdot |V| + \mu^2 \cdot |E|)\right)$.

Constraints (47) develop into smaller number of clauses if compared with the original constraints (41) which they replace because of the shorter right hand side in the implication in $F_{SIM}(\eta, \Sigma)$. Concretely, each implication from (47) develops into exactly 2 ternary clauses which gives $2 \cdot \eta \cdot \mu \cdot |E|$ ternary clauses in total.

Interconnection of auxiliary variables $\mathcal{E}_u^l$ with their meaning requires $\mu$ binary clauses per one implication from constraint (46). There are as many as $(\eta + 1) \cdot |V|$ such interconnections which results in $(\eta + 1) \cdot \mu \cdot |V|$ in total. Hence, the total number of clauses in $F_{SIM}(\eta, \Sigma)$ is $(\eta + 1) \cdot \left(\mu \cdot \left(\binom{|V|}{2} + 1\right) + |V| \cdot \binom{\mu}{2}\right) + \eta \cdot \mu \cdot (|V| + 2 \cdot |E|)$ which is $\mathcal{O}\left(\eta \cdot (\mu \cdot |V|^2 + \mu^2 \cdot |V| + \mu \cdot |E|)\right)$. ∎

**Proposition 9** (*PATHS AND $F_{DIR}(\eta, \Sigma)/F_{SIM}(\eta, \Sigma)$ SATISFACTION*). *A set $\Pi = \{\pi_1, \pi_2, \ldots, \pi_\mu\}$ of non-overlapping vertex disjoint paths in $\mathrm{Exp}_T(G, \eta)$ so that $\pi_i$ connects $[\alpha_0(a_i), 0]$ with $[\alpha_+(a_i), \eta]$ for $i = 1, 2, \ldots, \mu$ exists if and only if $F_{DIR}(\eta, \Sigma)$ is satisfiable. Moreover, paths $\pi_1, \pi_2, \ldots, \pi_\mu$ can be unambiguously constructed from satisfying valuation of $F_{DIR}(\eta, \Sigma)$ and vice versa. The same hold for $F_{SIM}(\eta, \Sigma)$.* ∎

**Sketch of proof.** Assume that non-overlapping vertex disjoint paths $\pi_1, \pi_2, \ldots, \pi_\mu$ exist so that $\pi_i$ connects $[\alpha_0(a_i), 0]$ with $[\alpha_+(a_i), \eta]$ in $\mathrm{Exp}_T(G, \eta)$. We will construct a satisfying valuation of $F_{DIR}(\eta, \Sigma)$ from $\pi_1, \pi_2, \ldots, \pi_\mu$.

Let $\pi_i = ([u_0, 0], [u_1, 1], [u_2, 2], \ldots, [u_\eta, \eta])$, with $u_l \in V$ for $l = 0, 1, \ldots, \eta$ where $u_0 = \alpha_0(a_i)$ and $u_\eta = \alpha_+(a_i)$, then variables $\mathcal{X}_{a_i, u_0}^0, \mathcal{X}_{a_i, u_1}^1, \ldots, \mathcal{X}_{a_i, u_\eta}^\eta$ will be set to *TRUE*. This setup of $\mathcal{X}_{a,v}^l$ variables will be set for every $i = 1, 2, \ldots, \mu$.

It is now easy to verify that all the constraints from the DIRECT encoding hold. Constraints (38) hold because each directed path $\pi_i$ intersects the time layer in exactly one vertex. Constraints (39) hold since directed paths are vertex disjoint. As paths go from one time layer to the next, constraints (40) hold as well. Finally, since paths are non-overlapping constraints (41) also hold.

Satisfying valuation of $F_{SIM}(\eta, \Sigma)$ requires assigning truth values to $\mathcal{E}_u^l$ variables in addition. Nevertheless, truth values of $\mathcal{E}_u^l$ are directly implied from assignment of truth values to $\mathcal{X}_{a,v}^l$ through constraints (46). Satisfaction of constraints (47) is ensured by satisfaction of constraints (41) and by transitivity of implication through the auxiliary $\mathcal{E}_u^l$. Connecting initial positions of agents with their goals by paths ensures satisfaction of constraints (42)-

(45) enforcing that initial time layer and the final time layer correspond to initial and goal arrangements of agent respectively.

If on the other hand we have satisfying valuation of $F_{DIR}(\eta, \Sigma)$, non-overlapping vertex disjoint paths can be constructed from it. Paths $\pi_1, \pi_2, \ldots, \pi_\mu$ will be constructed by following variables $\mathcal{X}_{a,v}^l$. Let $\pi_i = ([u_0, 0], [u_1, 1], [u_2, 2], \ldots, [u_\eta, \eta])$ where $\mathcal{X}_{a_i,u_0}^0, \mathcal{X}_{a_i,u_1}^1, \ldots, \mathcal{X}_{a_i,u_\eta}^\eta$ are $TRUE$. Single path is correctly defined as in each time layer $l \in \{0,1,\ldots,\eta\}$ and agent $a_i \in A$ there is exactly one $\mathcal{X}_{a_i,v}^l$ with $v \in V$ set to $TRUE$ which is ensured by constraints (38). Consecutive vertices in the path are connected by arcs which is ensured by constraints (40). If we consider all the paths together, then constraints (39) enforce that paths never intersects because two distinct agents cannot share a vertex. Finally, non-overlapping is ensured by constraints (41) since whenever non-trivial traversal between two consecutive time layers is made, no other agent can be involved in affected vertices. ∎

Again, recall that non-overlapping vertex disjoint paths correspond to CPF solutions (Proposition 1) which together with just proven result gives the following theorem.

**Theorem 4** (SOLUTION OF $\Sigma$ AND $F_{DIR}(\eta, \Sigma)/F_{SIM}(\eta, \Sigma)$ SATISFACTION). *A solution of a CPF $\Sigma = (G, A, \alpha_0, \alpha_+)$ exists if and only if there exist $\eta \in \mathbb{N}$ for that formula $F_{DIR}(\eta, \Sigma)$ is satisfiable. The same result holds for the simplified formula $F_{SIM}(\eta, \Sigma)$.* ∎

### 4.3. Summary of the Size Complexity of Propositional Encodings

Theoretical analysis of the size of encodings has been fine grained so far and it is not straightforward to see immediately how individual encodings compare with each other just by looking on expressions. Therefore, the extreme cases of all the expressions showing the number of variables and clauses have been evaluated and are shown here to provide a more complete picture.

The extreme cases concern the number of agents and neighborhood size in the graph $G$, which is either considered to be constant or asymptotically the same as the number of vertices.

Assumptions that the number of agents $\mu$ and the size of neighborhood in the graph asymptotically compares the number of vertices has been adopted in the space analysis in order to show the number of variables and clauses as much as possible in terms of the size of the input graph $G = (V, E)$.

Thus, we have following 4 scenarios (2 cases for each of 2 parameters):

- **Scenario (i):** The number of agents $\mu$ and the size of the neighborhood in $G$ is asymptotically the same as the number of vertices.
  (that is, $\mu \in \Theta(|V|)$ and $\deg_G(v) \in \Theta(|V|)$ for $\forall v \in V$).
  The assumption tells that the graph is highly occupied by agents and that the graph contains many edges. The consequence of the second assumption is also that the number of edges in the graph is asymptotically quadratic with respect to the number of vertices; that is, $|E| \in \Theta(|V|^2)$.

Space complexities in terms of the number of variables and clauses based upon above assumptions for this scenario are shown in Table 1.

**Table 1.** *Comparison of the Size Complexities of CPF Encodings – **Scenario (i)**.* The number of agents $\mu$ in this scenario is asymptotically the same as the number of vertices of $G$ (that is, $\mu \in \Theta(|V|)$) and the size of the vertex neighborhood in $G$ is also asymptotically the same as the number of vertices (that is, $\deg_G(v) \in \Theta(|V|)$ for $\forall v \in V$). For reference fine-grained complexity expression are shown as well.

| | #Variables<br>fine-grained/scenario (i) | #Clauses<br>fine-grained/scenario (i) |
|---|---|---|
| **INVERSE**<br>$F_{INV}(\eta, \Sigma)$ | $\mathcal{O}(\eta \cdot (|V| \cdot \lceil \log_2(\mu) \rceil + \sum_{v \in V} \lceil \log_2(\deg_G(v)) \rceil + |E|))$ | $\mathcal{O}(\eta \cdot (|V| \cdot \lceil \log_2(\mu) \rceil + |E| \cdot \lceil \log_2(\mu) \rceil + \sum_{v \in V} \deg_G(v) \cdot (\lceil \log_2(\deg_G(v)) \rceil)))$ |
| | $\mathcal{O}(\eta \cdot |V|^2)$ | $\mathcal{O}(\eta \cdot |V|^2 \cdot \lceil \log_2 |V| \rceil)$ |
| **ALL-DIFFERENT**<br>$F_{DIFF}(\eta, \Sigma)$ | $\mathcal{O}(\eta \cdot \mu \cdot |V|)$ | $\mathcal{O}\left(\eta \cdot \lceil \log_2 |V| \rceil \cdot \left(\binom{\mu}{2} + \mu \cdot |V|\right)\right)$ |
| | $\mathcal{O}(\eta \cdot |V|^2)$ | $\mathcal{O}(\eta \cdot |V|^2 \cdot \lceil \log_2 |V| \rceil)$ |
| **MATCHING**<br>$F_{MATCH}(\eta, \Sigma)$ | $\mathcal{O}(\eta \cdot (|E| + |V| \cdot \lceil \log_2(\mu) \rceil))$ | $\mathcal{O}\left(\eta \cdot \left((|V| + |E|) \cdot \lceil \log_2(\mu) \rceil + \sum_{v \in V} \binom{\deg_G(v)}{2}\right)\right)$ |
| | $\mathcal{O}(\eta \cdot |V|^2)$ | $\mathcal{O}(\eta \cdot |V|^3)$ |
| **DIRECT**<br>$F_{DIR}(\eta, \Sigma)$ | $\mathcal{O}(\eta \cdot \mu \cdot |V|)$ | $\mathcal{O}(\eta \cdot (\mu \cdot |V|^2 + \mu^2 \cdot |V| + \mu^2 \cdot |E|))$ |
| | $\mathcal{O}(\eta \cdot |V|^2)$ | $\mathcal{O}(\eta \cdot |V|^4)$ |
| **SIMPLIFIED**<br>$F_{SIM}(\eta, \Sigma)$ | $\mathcal{O}(\eta \cdot \mu \cdot |V|)$ | $\mathcal{O}(\eta \cdot (\mu \cdot |V|^2 + \mu^2 \cdot |V| + \mu \cdot |E|))$ |
| | $\mathcal{O}(\eta \cdot |V|^2)$ | $\mathcal{O}(\eta \cdot |V|^3)$ |

- **Scenario (ii):** The number of agents $\mu$ is asymptotically the same as the number of vertices while the size of the neighborhood in $G$ is asymptotically constant.
  (that is, $\mu \in \Theta(|V|)$ and $\deg_G(v) \in \Theta(1)$ for $\forall v \in V$).

The second assumption tells the graph is sparse and can be intuitively compared to *planar graphs* [36] that are very common in practice. The assumption also tells that the number of edges is asymptotically the same as the number of vertices; that is, $|E| \in \Theta(|V|)$.

Space complexities for this scenario are shown in Table 2.

**Table 2.** *Comparison of the size complexities of CPF encodings – **Scenario (ii)**.* The number of agents $\mu$ in this scenario is asymptotically the same as the number of vertices of $G$ (that is, $\mu \in \Theta(|V|)$) while the size of the vertex neighborhood in $G$ is asymptotically constant (that is, $\deg_G(v) \in \Theta(1)$ for $\forall v \in V$).

|  | #Variables scenario (ii) | #Clauses scenario (ii) |
|---|---|---|
| INVERSE $F_{INV}(\eta, \Sigma)$ | $\mathcal{O}(\eta \cdot |V| \cdot \lceil \log_2 |V| \rceil)$ | $\mathcal{O}(\eta \cdot |V| \cdot \lceil \log_2 |V| \rceil)$ |
| ALL-DIFFERENT $F_{DIFF}(\eta, \Sigma)$ | $\mathcal{O}(\eta \cdot |V|^2)$ | $\mathcal{O}(\eta \cdot |V|^2 \cdot \lceil \log_2 |V| \rceil)$ |
| MATCHING $F_{MATCH}(\eta, \Sigma)$ | $\mathcal{O}(\eta \cdot |V| \cdot \lceil \log_2 |V| \rceil)$ | $\mathcal{O}(\eta \cdot |V| \cdot \lceil \log_2 |V| \rceil)$ |
| DIRECT $F_{DIR}(\eta, \Sigma)$ | $\mathcal{O}(\eta \cdot |V|^2)$ | $\mathcal{O}(\eta \cdot |V|^3)$ |
| SIMPLIFIED $F_{SIM}(\eta, \Sigma)$ | $\mathcal{O}(\eta \cdot |V|^2)$ | $\mathcal{O}(\eta \cdot |V|^3)$ |

- **Scenario (iii):** The number of agents $\mu$ is asymptotically constant while the size of the neighborhood in $G$ is asymptotically the same as the number of vertices.
  (that is, $\mu \in \Theta(1)$ and $\deg_G(v) \in \Theta(|V|)$ for $\forall v \in V$).
  This scenario can be intuitively regarded as a planar graph densely occupied by agents. Space complexities for this scenario are shown in Table 3.

- **Scenario (iv):** The number of agents $\mu$ and the size of the neighborhood in $G$ are both asymptotically constant.
  (that is, $\mu \in \Theta(1)$ and $\deg_G(v) \in \Theta(1)$ for $\forall v \in V$).
  Again, this scenario can be intuitively regarded as a planar graph with few agents inside. Space complexities for this scenario are shown in Table 4.

The measure used here for comparison of encodings is that the smaller number of variables or clauses the better. This is usually a realistic measure as the search space often correlates with the number of (decision) variables when solving the propositional formula satisfiability problem by standard search procedures. Similarly, the small number of clauses means that the overall size of the propositional formula is small and thus it is easier for the overall processing. Nevertheless, such small formula preference should be considered just as an intuitive measure since sometimes lot of variables may be derivable from values of other variables (thus they do not increase size of the search space) and sometimes more clauses may improve propagation.

**Table 3.** *Comparison of the size complexities of CPF encodings – **Scenario (iii)**. The number of agents $\mu$ in this scenario is constant (that is, $\mu \in \Theta(1)$) while the size of the vertex neighborhood is asymptotically the same as the number of vertices of $G$ (that is, $\deg_G(v) \in \Theta(|V|)$ for $\forall v \in V$).*

|  | #Variables<br>scenario (iii) | #Clauses<br>scenario (iii) |
|---|---|---|
| INVERSE<br>$F_{INV}(\eta, \Sigma)$ | $\mathcal{O}(\eta \cdot |V|^2)$ | $\mathcal{O}(\eta \cdot |V|^2 \cdot \lceil \log_2 |V| \rceil)$ |
| ALL-DIFFERENT<br>$F_{DIFF}(\eta, \Sigma)$ | $\mathcal{O}(\eta \cdot |V|)$ | $\mathcal{O}(\eta \cdot |V| \cdot \lceil \log_2 |V| \rceil)$ |
| MATCHING<br>$F_{MATCH}(\eta, \Sigma)$ | $\mathcal{O}(\eta \cdot |V|^2)$ | $\mathcal{O}(\eta \cdot |V|^3)$ |
| DIRECT<br>$F_{DIR}(\eta, \Sigma)$ | $\mathcal{O}(\eta \cdot |V|)$ | $\mathcal{O}(\eta \cdot |V|^2)$ |
| SIMPLIFIED<br>$F_{SIM}(\eta, \Sigma)$ | $\mathcal{O}(\eta \cdot |V|)$ | $\mathcal{O}(\eta \cdot |V|^2)$ |

Several conclusions can be made upon asymptotic numbers of variables and clauses in individual encodings presented in Table 1 - Table 4. In cases with many agents and dense graphs (corresponding to scenario (i)), INVERSE and ALL-DIFFERENT encodings excel in a small number of clauses.

**Table 4.** *Comparison of the size complexities of CPF encodings – **Scenario (iv)**. Both the number of agents $\mu$ as well as the size of the vertex neighborhood are asymptotically constant in this scenario (that is, $\mu \in \Theta(1)$ and $\deg_G(v) \in \Theta(1)$ for $\forall v \in V$).*

|  | #Variables<br>scenario (iv) | #Clauses<br>scenario (iv) |
|---|---|---|
| INVERSE<br>$F_{INV}(\eta, \Sigma)$ | $\mathcal{O}(\eta \cdot |V|)$ | $\mathcal{O}(\eta \cdot |V|)$ |
| ALL-DIFFERENT<br>$F_{DIFF}(\eta, \Sigma)$ | $\mathcal{O}(\eta \cdot |V|)$ | $\mathcal{O}(\eta \cdot |V| \cdot \lceil \log_2 |V| \rceil)$ |
| MATCHING<br>$F_{MATCH}(\eta, \Sigma)$ | $\mathcal{O}(\eta \cdot |V|)$ | $\mathcal{O}(\eta \cdot |V|)$ |
| DIRECT<br>$F_{DIR}(\eta, \Sigma)$ | $\mathcal{O}(\eta \cdot |V|)$ | $\mathcal{O}(\eta \cdot |V|^2)$ |
| SIMPLIFIED<br>$F_{SIM}(\eta, \Sigma)$ | $\mathcal{O}(\eta \cdot |V|)$ | $\mathcal{O}(\eta \cdot |V|^2)$ |

When we have many agents and relatively sparse graphs (scenario (ii)), which is the most common case in practice, then INVERSE and MATCHING encodings excel in both, in the number of variables as well as in the number of clauses.

The remaining two scenarios (scenario (iii) and (iv)) can be regarded as non-cooperative scenarios since the number of agents is constant and hence the interaction among them is limited. The ALL-DIFFERENT encoding is the most space saving in a case with dense graphs (scenario (iii)) while INVERSE and MATCHING encodings are the most space saving on sparse graphs (scenario (iv)).

Observe that DIRECT and SIMPLIFIED encodings do not excel in any of the suggested scenarios. This is mostly caused by the fact that no binary encoding of finite domain state variables, which significantly reduces the size of representation of the state variable using propositional variables, is used in these two encodings.

### 4.3.1. Knowledge Compilation – Distance Heuristics

Encodings based on time expansion graph can be be further enhanced by a so called *distance heuristic*. Intuitively said, a path indicating the trajectory of a given agent cannot go through vertices that there too far from the initial or the goal vertex under given time constraints. In other words, vertices at a given time layer where the distance to the initial position of the agent is larger than the time elapsed for the time layer (which equals to the position of the time layer in the time expansion graph) or where the distance to the goal vertex is larger than the time that remains for the given time layer (which equals to the position of the time layer in the time expansion graph counted from the end) can never be visited by the agent. The just described time consideration can be easily formalized in the time expansion graphs through existence of directed paths.

The knowledge of these impassable vertices can rule out the occurrence of the agent in them from further consideration during the search for a solution and consequently reduce the search space.

Assume a time expansion graph $\text{Exp}_T(G,\eta)$ for CPF $\Sigma = (G, A, \alpha_0, \alpha_+)$; let $\text{dist}_D^{\rightarrow}(u,v)$ denote the length of the shortest directed path connecting $u$ to $v$ in a given digraph $D = (X, F)$; $\text{dist}_D^{\rightarrow}(u,v) = \infty$ if there is no path connecting $u$ to $v$ in $D$.

**Proposition 10** (*DISTANCE HEURISTIC*). *Any solution $\vec{s} = [\alpha_0, \alpha_1, \alpha_2, \dots, \alpha_\eta]$ to $\Sigma$ satisfies that* $\text{dist}_{\text{Exp}_T(G,\eta)}^{\rightarrow}([\alpha_0(a_i), 0], [\alpha_l(a_i), l]) < \infty$ *and* $\text{dist}_{\text{Exp}_T(G,\eta)}^{\rightarrow}([\alpha_l(a_i), l], [\alpha_\eta(a_i), \eta]) < \infty$ *for every $i \in \{1,2,\dots,\mu\}$ and $l \in \{0,1,\dots,\eta\}$.* ∎

**Proof.** The proposition is in fact a direct consequence of Proposition 1. If $\text{dist}_{\text{Exp}_T(G,\eta)}^{\rightarrow}([\alpha_0(a_i), 0], [v, l]) = \infty$ or $\text{dist}_{\text{Exp}_T(G,\eta)}^{\rightarrow}([v, l], [\alpha_\eta(a_i), \eta]) = \infty$ for some $v \in V$ then there is no directed path connecting $[\alpha_0(a_i), 0]$ and $[\alpha_\eta(a_i), \eta]$ going through $[v, l]$. A fortiori, there is no path connecting $[\alpha_0(a_i), 0]$ and $[\alpha_\eta(a_i), \eta]$ visiting $[v, l]$ that does not overlap and does not intersect other paths. Hence, $\alpha_l(a_i) \neq v$. ∎

The above proposition can be used to design a heuristic. All the vertices $[v_j, l]$ with $v_j \in V$ and $l \in \{0,1,\ldots,\eta\}$ in $\text{Exp}_T(G, \eta)$ for that $\text{dist}^{\rightarrow}_{\text{Exp}_T(G,\eta)}([\alpha_0(a_i), 0], [v_j, l]) = \infty$ or $\text{dist}^{\rightarrow}_{\text{Exp}_T(G,\eta)}([v_j, l], [\alpha_\eta(a_i), \eta]) = \infty$ can be excluded from trajectories corresponding to agent $a_i$ (in original graph it translates to the requirement that agent $a_i$ cannot enter $v_j$ at time step $l$). In all the encodings, this can be done easily as follows:

$$\mathcal{A}^l_{v_j} \neq i \quad \text{in the INVERSE and MATCHING encoding} \tag{30}$$

$$\mathcal{L}^l_{a_i} \neq j \quad \text{in the ALL-DIFFERENT encoding} \tag{31}$$

$$\neg \mathcal{X}^l_{a_i, v_j} \quad \text{in the DIRECT/SIMPLIFIED encoding} \tag{32}$$

The inequality between a bit vector and a constant is encoded as a single clause that forbids the bit vector to take that constant. That is, at least one bit must disagree with binary representation of the constant. For example, the inequality $\mathcal{A}^l_v \neq c$ is encoded as follows:

$$\text{con}_{\neq}(\mathcal{A}^l_v, c) = \bigvee_{\hat{\imath}=0}^{\lceil \log_2 \mu \rceil - 1} \neg \text{lit}(\mathcal{A}^l_v, c, \hat{\imath}) \tag{33}$$

It holds that added inequalities are logical consequences of the encoded propositional formulae (that is, for example $F_{INV}(\eta, \Sigma) \Rightarrow \mathcal{A}^l_{v_j} \neq i$ is a valid formula). Thus in theory, the SAT solver should be able to infer that some vertices are not reachable at certain time steps. However, it may be costly to derive such a fact for the SAT solver while the same knowledge can be obtained easily in advance and compiled directly into the formula almost without any increase of its complexity.

## 5. Experimental Evaluation

Experimental evaluation has been focused on measuring the actual size of suggested encodings and on measuring runtime when encodings are used for makespan optimal CPF solving.

The solving procedure presented as Algorithm 1 was used as a core framework for our makespan optimal CPF solving technique (that is, the sequential increasing strategy for querying the SAT solver was used) while suggested individual propositional encodings can be regarded as its exchangeable modules. The SAT solver itself was connected to the solving technique as another external module. All the implemented encodings used build-in distance heuristic discussed in section 4.3.1.

The SAT-based CPF solving procedure was implemented in C++ as well as procedures for generating propositional formulae from given CPF and makespan bound (solving procedure and formulae generation were compiled together as a single executable program).

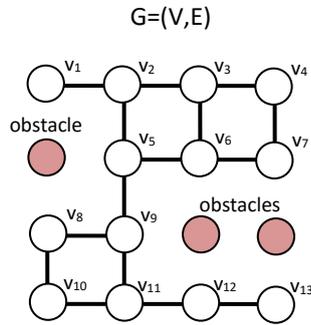

**Figure 6.** *Four-connected grid of size 4×4 with 3 obstacles.* Positions of obstacles within the grid are depicted though they are actually not present in the graph.

We used `glucose 3.0` SAT solver [2] in our tests, which is justified by the fact that this SAT solver ranked among the winners in recent SAT Competitions [3] in the category of hard combinatorial problems to which we consider CPF belongs as well. The SAT solver was a separate module and was called externally by the CPF solving procedure (the solving procedure always generated a formula into a file, which was then submitted as input to the SAT solver; the answer of the SAT solver was generated into another file, from which the procedure read it and further processed).

### 5.1. Benchmark Setup

We followed benchmark setup suggested by Silver in [20]. Four-connected grids of various sizes were used to model environments in testing instances. The size of grids ranged from 6×6 to 12×12 with 20% of randomly selected vertices occupied by obstacles (obstacle was represented by a missing vertex in grid – see Figure 6). Initial and goal arrangements of agents was random – the random arrangement of agents has been obtained by placing agents one by one while the position has been uniformly randomly picked from the remaining unoccupied vertices. Only solvable instances were taken into runtime tests.

To allow full reproducibility of presented results all the source codes and experimental data were posted on-line on: http://ktiml.mff.cuni.cz/~surynek/research/j-encoding-2015.

### 5.2. Encoding Size Evaluation

The size of propositional formulae was tested for discussed 4-connected grids with the increasing number of agents inside. The number of agents ranged from 1 up to the half all the vertices in the graph.

For each number of agents, 10 random CPF instances were generated and their characteristics were measured. Formulae corresponding to all the suggested encodings were generated for each number of agents. The number of layers in time expansion graphs was fixed and set relatively to the size of the instance – it was 12 for 6×6 grid; 16 for 8×8 grid; and 24 for 12×12 grid.

**Table 5.** *Size comparison of propositional encodings of CPF over 6×6 grid.* CPF instances are generated over the 4-connected grid of size 6×6 with 20% of vertices occupied by obstacles. The number of time layers in corresponding time expansion graph $\eta$ is 12. The number of *variables* and *clauses*, the *ratio* of the number of clauses and the number of variables, and the *average clause length* are listed for different sizes of the set of agents *A*. Small size of the formula and short clauses (they support unit propagation) are preferred – best values for each measure according to this preference are shown in bold. DIRECT and SIMPLIFIED encodings are best in number of measures on 6×6 grid.

| Grid 6×6 |  |  | INVERSE | | ALL-DIFFERENT | | MATCHING | | DIRECT | | SIMPLIFIED | |
|---|---|---|---|---|---|---|---|---|---|---|---|---|
| \|Agents\| | | | | | | | | | | | | |
| 1 | #Variables | Ratio | 3 384.3 | 3.692 | 701.4 | 4.506 | 1 841.1 | 5.595 | **342.0** | 17.685 | 684.0 | 2.192 |
|   | #Clauses | Length | 12 494.3 | 2.622 | 3 160.7 | 2.979 | 10 300.6 | 2.436 | 6 048.2 | **2.261** | **1 499.6** | 2.587 |
| 2 | | | 3 738.3 | 4.551 | 1 723.5 | 4.173 | 2 195.1 | 6.149 | **684.0** | 20.725 | 1 026.0 | 3.354 |
|   | | | 17 012.0 | 2.599 | 7 191.7 | 2.980 | 13 497.1 | 2.512 | 14 176.0 | 2.353 | **3 441.4** | 2.562 |
| 4 | | | 4 092.3 | 5.403 | 4 127.5 | 3.729 | 2 549.1 | 6.777 | **1 368.0** | 25.558 | 1 710.0 | 4.653 |
|   | | | 22 110.2 | 2.642 | 15 392.6 | 3.026 | 17 274.1 | 2.632 | 34 962.7 | 2.427 | **7 956.1** | **2.423** |
| 8 | | | 4 446.3 | 6.348 | 12 066.7 | 3.250 | 2 903.1 | 7.601 | **2 736.0** | 36.324 | 3 078.0 | 6.964 |
|   | | | 28 225.0 | 2.794 | 39 216.1 | 3.060 | 22 067.7 | 2.867 | 99 381.3 | 2.543 | **21 436.3** | **2.319** |
| 16 | | | 4 800.3 | 7.609 | 38 791.0 | 2.830 | **3 257.1** | 8.919 | 5 472.0 | 56.375 | 5 814.0 | 11.062 |
|   | | | 36 527.1 | 3.133 | 109 781.6 | 3.104 | **29 048.6** | 3.313 | 308 484.7 | 2.633 | 64 314.8 | **2.201** |

The average number of propositional variables, average number of clauses, ratio between the number of clauses and variables, and average clause length were calculated for each encoding and number of agents out of 10 randomly generated instances. Partial results are shown in Table 5, Table 6, and Table 7 – preferred values of individual characteristics are listed in bold.

The number of variables and clauses directly correspond to the size of formulae. Preference is given to formulae that are smaller as they are expected to be easier to solve as well as easier for processing.

The ratio of the number of clauses and the number of variables is an important measure of the difficulty of propositional formula. Formulae that are *under-constrained* or *over-constrained* are easier to solve [7] (easily satisfiable or easily unsatisfiable respectively) and hence such situation is preferred in formulae encoding CPF.

A very important characteristic is the average length of clause while short clauses are preferred since they support *unit propagation* [7], which allows deriving values for other variable without search.

Results indicate that DIRECT and SIMPLIFIED encodings have best size characteristics with respect to the small size preference in cases with small number of agents in the instance. This result can be observed for all the sizes of the grid modeling the environment. As the neighborhood connectivity in 4-connected grids can be regarded as constant; that is, $\deg_G(v) \in \Theta(1)$ for $\forall v \in V$, the cases, where DIRECT and SIMPLIFIED encoding have best size characteristics, roughly correspond to scenario (iv). However, theoretical asymptotic formula size estimations suggest different results - DIRECT and SIMPLIFIED encodings should be same as other encodings in terms of the number of variables and worse than other encodings in terms of the number of variables. Hence, experimental evaluation has shown a surprising result in this aspect.

**Table 6.** *Size comparison of propositional encodings of CPF over 8×8 grid.* The number of time layers in the corresponding time expansion graph is 16. DIRECT and SIMPLIFIED encodings have fewer variables and clauses for small number of agents while MATCHING encoding is better in these measures for many agents in the instance.

| Grid 8×8 |  |  | INVERSE |  | ALL-DIFFERENT |  | MATCHING |  | DIRECT |  | SIMPLIFIED |  |
|---|---|---|---|---|---|---|---|---|---|---|---|---|
| \|Agents\| |  |  |  |  |  |  |  |  |  |  |  |  |
| 1 | #Variables | *Ratio* | 4 520.3 | *3.748* | 1 489.3 | *5.325* | 4 520.3 | *5.710* | **814.4** | *28.539* | 1 628.8 | *2.078* |
|   | #Clauses | *Length* | 25 881.1 | *2.616* | 7 930.4 | *3.057* | 25 881.1 | *2.441* | 23 241.9 | *2.149* | **3 384.6** | *2.550* |
| 4 |  |  | 10 019.5 | *5.532* | 7 834.5 | *4.440* | 6 181.1 | *6.984* | **3 257.6** | *35.589* | 4 072.0 | *4.420* |
|   |  |  | 55 437.0 | *2.641* | 34 781.9 | *3.103* | 43 171.0 | *2.640* | 115 934.3 | *2.272* | **17 997.8** | *2.374* |
| 8 |  |  | 10 849.9 | *6.519* | 21 875.4 | *3.831* | 7 011.5 | *7.851* | **6 515.2** | *45.635* | 7 329.6 | *5.736* |
|   |  |  | 70 725.9 | *2.792* | 83 794.2 | *3.113* | 55 050.3 | *2.874* | 297 319.9 | *2.390* | **49 381.3** | *2.694* |
| 16 |  |  | 11 680.3 | *7.820* | 67 088.3 | *3.231* | **7 841.9** | *9.215* | 13 030.4 | *64.506* | 13 844.8 | *10.853* |
|   |  |  | 91 344.5 | *3.127* | 216 745.4 | *3.147* | **72 259.3** | *3.315* | 840 540.6 | *2.505* | 150 259.2 | *2.180* |
| 32 |  |  | 12 510.7 | *9.765* | 230 753.0 | *2.802* | **8 672.3** | *11.494* | 26 060.8 | *105.084* | 26 875.2 | *19.002* |
|   |  |  | 122 170.3 | *3.733* | 646 616.2 | *3.168* | **99 675.5** | *4.045* | 2 738 584.7 | *2.621* | 510 672.1 | *2.111* |

If the number of agents is higher, the MATCHING encoding dominates in the size characteristics for all the size of the grid. It has the fewest number of propositional variables as well as the fewest number of clauses. If we consider that this case roughly correspond to scenario (ii), these observations correspond to theoretical asymptotic estimations, which indicate that MATCHING encoding together with INVERSE encoding should be smallest (note, that the INVERSE encoding is the second smallest according to experimental results).

**Table 7.** *Size comparison of propositional encodings of CPF over 12×12 grid.* The number of time layers in the time expansion graph is 24 here. The MATCHING encoding is clearly the smallest encoding for larger number of agents.

| Grid 12×12 |  |  | INVERSE |  | ALL-DIFFERENT |  | MATCHING |  | DIRECT |  | SIMPLIFIED |  |
|---|---|---|---|---|---|---|---|---|---|---|---|---|
| \|Agents\| |  |  |  |  |  |  |  |  |  |  |  |  |
| 1 | #Variables | *Ratio* | 29 798.7 | *3.903* | 4 973.9 | *6.218* | 15 961.3 | *5.927* | **2 767.2** | *60.721* | 5 534.4 | *2.094* |
|   | #Clauses | *Length* | 116 302.8 | *2.635* | 30 928.8 | *3.031* | 94 603.2 | *2.443* | 168 027.8 | *2.073* | **11 587.0** | *2.578* |
| 8 |  |  | 38 172.3 | *6.752* | 55 602.1 | *4.887* | 24 334.9 | *8.130* | **22 137.6** | *77.789* | 24 904.8 | *6.707* |
|   |  |  | 257 739.9 | *2.793* | 271 730.3 | *3.088* | 197 835.9 | *2.871* | 1 722 059.3 | *2.230* | **167 026.1** | *2.289* |
| 16 |  |  | 40 963.5 | *8.062* | 153 047.5 | *4.290* | **27 126.1** | *9.510* | 44 275.2 | *97.349* | 47 042.4 | *11.540* |
|   |  |  | 330 249.1 | *3.115* | 656 615.4 | *2.999* | **257 974.6** | *3.300* | 4 310 137.7 | *2.343* | 542 862.4 | *2.059* |
| 32 |  |  | 43 754.7 | *10.049* | 475 135.0 | *3.428* | **29 917.3** | *11.843* | 88 550.4 | *136.888* | 91 317.6 | *18.953* |
|   |  |  | 439 680.0 | *3.701* | 1 628 634.8 | *3.148* | **354 306.4** | *4.021* | 12 121 528.6 | *2.475* | 1 730 745.7 | *2.112* |
| 64 |  |  | 46 545.9 | *13.340* | 1 626 205.9 | *2.898* | **32 708.5** | *15.985* | 177 100.8 | *216.610* | 179 868.0 | *35.0127* |
|   |  |  | 620 942.7 | *4.632* | 4 713 520.1 | *3.183* | **522 834.3** | *5.065* | 38 361 723.7 | *2.594* | 6 297 660.9 | *2.062* |

DIRECT and SIMPLIFIED encodings excel in terms of the ratio of the number of clauses to the number of variables. Both encodings tend to be over-constrained, which intuitively suggest easier proving of unsatisfiability. The average length of clauses is shortest for the SIMPLIFIED encoding. As the number of agents increases the average clause length converges towards 2 for all the sizes of the grid (that is, most of clauses are binary in the SIMPLIFIED encoding).

Above observations of static characteristics of encodings indicate that MATCHING encoding and especially SIMPLIFIED encoding should perform well in CPF solving (or at least better than other encodings).

## 5.3. Runtime Evaluation

We re-implemented A*-based OD+ID CPF solving procedure [23] in C++ with the objective function for minimizing the makespan and compared it with our SAT based solving method in order to provide broader picture regarding the runtime evaluation.

Again, CPFs over 4-connected grids of sizes 6×6, 8×8, and 12×12 with 20% of vertices occupied by randomly placed obstacles were used. Initial and goal arrangements of agents were generated randomly. Runtime evaluation was done for the increasing number of agents in instances while for each number of agents 10 random instances were generated and solved. All the instances used for evaluation were solvable.

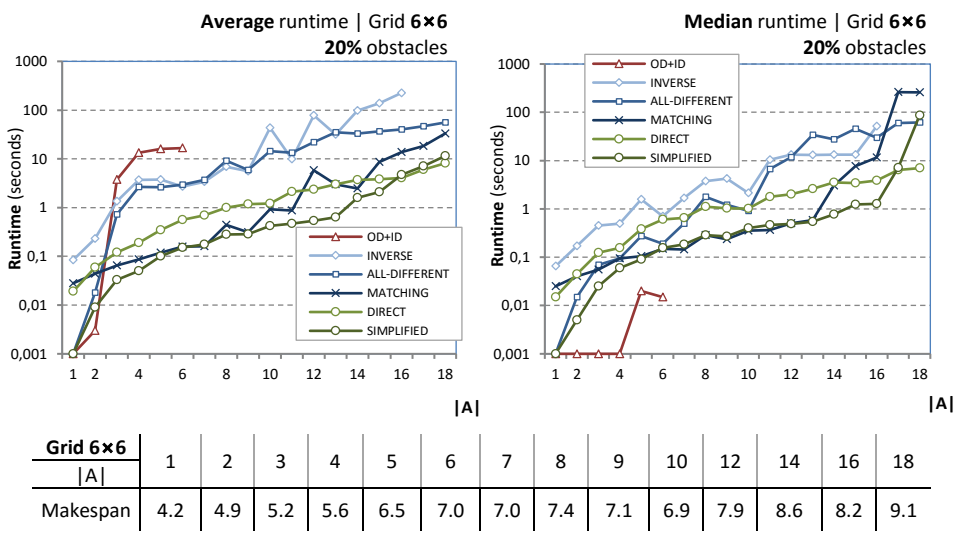

| Grid 6×6 |A| | 1 | 2 | 3 | 4 | 5 | 6 | 7 | 8 | 9 | 10 | 12 | 14 | 16 | 18 |
|---|---|---|---|---|---|---|---|---|---|---|---|---|---|---|
| Makespan | 4.2 | 4.9 | 5.2 | 5.6 | 6.5 | 7.0 | 7.0 | 7.4 | 7.1 | 6.9 | 7.9 | 8.6 | 8.2 | 9.1 |

**Figure 7.** *Runtime evaluation over 6×6 4-connected grid with 20% of vertices occupied by obstacles.* A*-based method OD+ID and SAT-based method with INVERSE, ALL-DIFFERENT, MATCHING, DIRECT, and SIMPLIFIED encodings are compared on random CPF instances on the grid. Average and median runtimes out of runtimes on 10 random instances are shown; average optimal makespans are also shown. OD+ID method does not scale for higher number of agents while SAT-based solving performs better with many agents. Particularly SIMPLIFIED encoding performs as best. Up to two orders of magnitude are between the best and worst encoding in runtime.

The timeout for single CPF instance solving was set to 256 seconds (approximately 4 minutes). The number of agents was increased until all the 10 random instances were solvable within the given timeout – that is, each solving method (encoding) is characterized by the maximum number of agents for which it is able to solve all the 10 random instances within the given timeout.

The average and median runtimes were calculated out of these 10 instances for all the tested methods. In the case of SAT based CPF solving methods, the runtime is a sum of the runtime of the core CPF solving procedure (corresponding to Algorithm 1) plus runtimes of all the runs of the SAT solver invoked by the core procedure.

Runtime results together with average optimal makespan are shown in Figure 7, Figure 8, and Figure 9[1] (note that, all the methods generate solutions of the same optimal makespan).

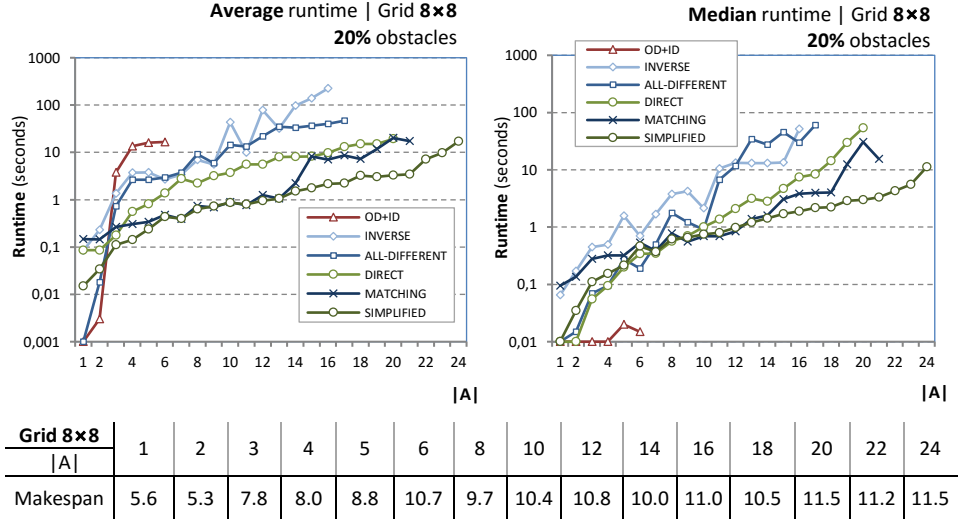

| Grid 8×8 |   |   |   |   |   |   |   |   |   |   |   |   |   |   |
|---|---|---|---|---|---|---|---|---|---|---|---|---|---|---|
| \|A\| | 1 | 2 | 3 | 4 | 5 | 6 | 8 | 10 | 12 | 14 | 16 | 18 | 20 | 22 | 24 |
| Makespan | 5.6 | 5.3 | 7.8 | 8.0 | 8.8 | 10.7 | 9.7 | 10.4 | 10.8 | 10.0 | 11.0 | 10.5 | 11.5 | 11.2 | 11.5 |

**Figure 8.** *Runtime evaluation over 8×8 4-connected grid with 20% of vertices occupied by obstacles.* Again, the SAT-based solving with SIMPLIFIED encoding performs as best for higher number of agents. The MATCHING encoding also starts with promising performance but it quickly degrades for more than approximately 14 agents.

It can be observed that OD+ID, although it is the fastest for small number of agents, does not scale up as the runtime quickly blows up for more agents. The SAT-based solving method with all the encodings performs better and scales up for higher number of agents. Particularly, the SIMPLIFIED encoding performs as best in all the sizes of the grid followed by MATCHING, DIRECT, ALL-DIFFERENT, and INVERSE encodings respectively.

Note that the good performance of the SIMPLIFIED encoding has been predicted by the static analysis of encodings (particularly, it has been assumed to support unit propagation well). Another well competing MATCHING encoding had been predicted to have a good performance as well due to its small size in testing instances.

An interesting behavior can be observed with MATCHING encoding that start with almost the same promising performance as the SIMPLIFIED encoding for small number of agents, but it quickly degrades and it is eventually outperformed by the DIRECT encoding on 6×6 and 12×12 grids for higher number of agents (in case of the 8×8 grid, the degradation of the MATCHING encoding can be observed as well but it is less significant – the DIRECT encoding reached the timeout before it could overtake the MATCHING encoding).

---

[1] All the runtime measurements were done on a machine with the 4-core CPU Xeon 2.0GHz and 12GB RAM under Linux kernel 3.5.0-48. Although we used multiple cores to run experiments in parallel, the individual instances were solved in a single thread (that is, the core solving procedure and all its call to the SAT solver were run in single thread).

Instances with occupancy by agents up to 62% are solvable within the given timeout in the 6×6 grid by using the SIMPLIFIED encoding. This figure is 46% for the 8×8 grid and 28% for the 12×12 grid for the SIMPLIFIED encoding. OD+ID method can solve instances with occupancy up to 24%, 13%, and 7% in 6×6, 8×8, 12×12 grids respectively. Thus, approximately 3 times more agents are solvable with SAT based method than with OD+ID in given testing instances.

The general conclusion from the above experimental evaluation is also that the binary encoding used for encoding finite domain state variables in the INVERSE, ALL-DIFFERENT, and MATCHING encoding contributes to the small size but it is questionable if it contributes the overall performance as these encodings clearly performed worse than the DIRECT and SIMPLIFIED encodings that did not rely on the binary encoding.

On the other hand, the simple design of the DIRECT and SIMPLIFIED encodings is not at the expense of the performance of their solving. The simple design of variables allowed modeling constraints using short clauses that significantly support intensive unit propagation, which is most likely the key factor for the good performance of both encodings – especially in the case of the SIMPLIFIED encoding.

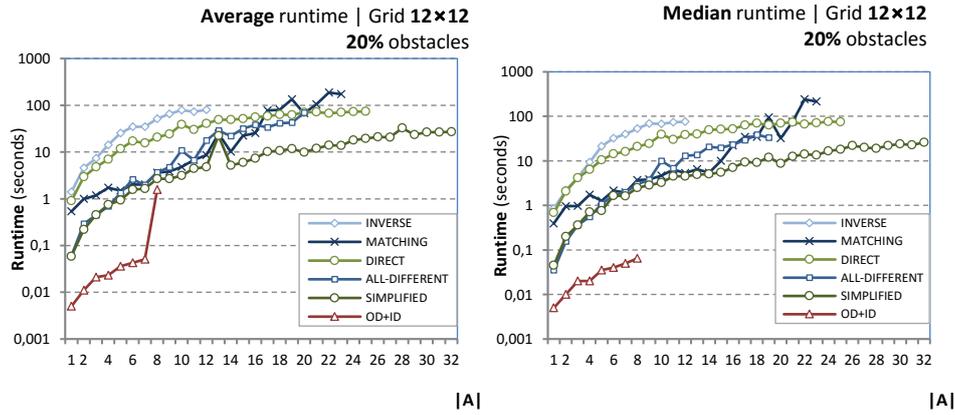

| Grid 12×12 |   |   |   |   |   |   |   |   |   |   |   |   |   |
|---|---|---|---|---|---|---|---|---|---|---|---|---|---|
| \|A\| | 1 | 2 | 3 | 4 | 6 | 8 | 10 | 12 | 16 | 18 | 20 | 24 | 28 | 32 |
| Makespan | 7.5 | 10.6 | 11.4 | 12.7 | 13.9 | 15.3 | 13.8 | 14.9 | 15.5 | 16.3 | 14.6 | 17.3 | 15.9 | 16.4 |

**Figure 9.** *Runtime evaluation over 12×12 4-connected grid with 20% of vertices occupied by obstacles.* The best performance is exhibited by the SIMPLIFIED encoding again. The MATCHING encoding is able to solve second highest number of agents in the given timeout.

### 5.4. Solution Quality Evaluation

Although all the solutions generated by the suggested SAT based solving techniques are makespan optimal, that is, the best with respect to our objective function, they may differ in other aspects. Particularly important is the *total number of moves* performed by agents (also called a *sum of costs*) which can be regarded as the total energy consumed by agents to perform their movements. The total number of moves is also considered as an objective function in several approaches to CPF solving such as [21, 22]. Hence, it is interesting what do solutions generated by makespan optimal SAT solving look like with respect to the total number of moves despite the fact that this aspect has been completely disregarded in the design of propositional encodings of CPF.

The way in which a given problem is encoded into propositional formula greatly affects heuristics the SAT solver uses for selecting variables and their values. Values selected to satisfy the formula are then reflected in the CPF solution reconstructed from its satisfying valuation. Although not a rule, SAT solvers in their default settings usually prefer assigning *FALSE* value if it is not more advantageous than to assign value *TRUE*.

Observe that only values assigned to visible propositional variables are directly reflected in the resulting CPF solution. Visible propositional variables in the suggested encodings are either part of a directly encoded state (DIRECT and SIMPLIFIED encodings) or part of a binary encoded bit vector (INVERSE, ALL-DIFFERENT, and MATCHING encodings).

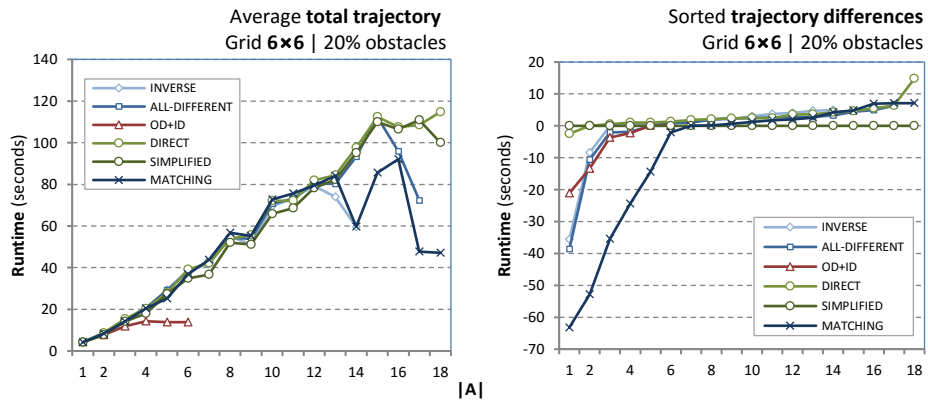

**Figure 10.** *Solution quality comparison over 6×6 4-connected grid.* The total number of moves in optimal solutions obtained by each tested method is compared for the growing number of agents in the grid (left part). Sorted differences in total number of moves from the number of moves generated with the SIMPLIFIED encoding are also shown (right part). The SIMPLIFIED encoding yields a solution with the fewer number of moves than other methods in about one third of all the generated solutions.

Propositional variables within directly encoded state directly correspond to occupancy of a vertex or an edge by a fixed agent. Assignment of value *FALSE* to a propositional variable of the directly encoded state corresponds to no occupancy by the given fixed agent. Complete no-occupancy appears if and only if all the propositional variables directly encoding the state are set to *FALSE* for all the agents.

The interpretation of bit vector propositional variables is that occupancy of a corresponding vertex or an edge appears if any of the propositional variables within the bit vector is set to $TRUE$. No occupancy corresponds to the assignment of integer zero to the bit vector, which means to assign $FALSE$ to all the propositional variables, which the bit vector consists of.

If we assume that the SAT solver tries to find a solution conservatively; that is, it prefers to assign $FALSE$ values, then it seems that using encodings with visible variables that directly encode states results in smaller vertex and edge occupancy, which correspond to CPF solutions consisting of fewer total number of moves.

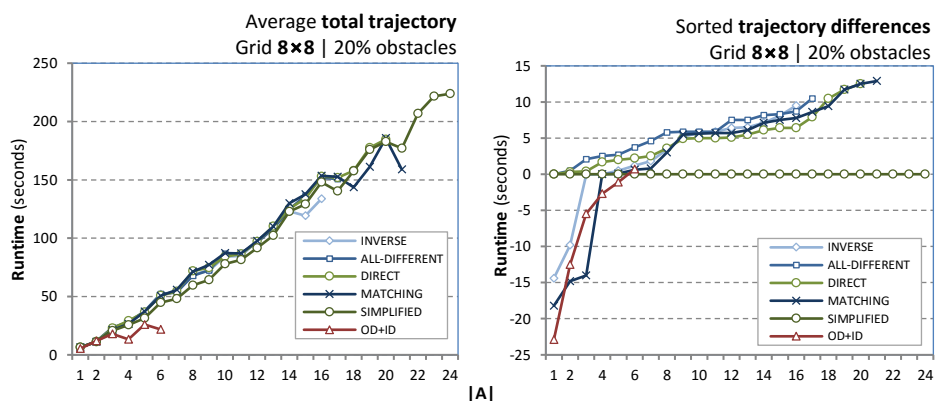

**Figure 11.** *Solution **quality comparison** over 8×8 4-connected grid.* The SIMPLIFIED encoding yields solutions with fewest moves in approximately 75% of cases of solution generation. Note that A*-based OD+ID generates solutions with even fewer moves but it does not scale up enough to show its qualities for higher density of agents.

The reasoning behind this hypothesis assumes to have a set of agents $A$ and a location (vertex/edge) that is to be occupied by at most one agent from $A$. The occupancy of the location is modeled by directly encoded state in one scenario and as a binary encoded bit vector in the second scenario. The first scenario yields $|A|$ propositional variables with $|A| + 1$ allowed assignments - one of these assignments corresponding to no occupancy of the location assigns $FALSE$ to all the propositional variables; other allowed assignments have just a single propositional variable set to $TRUE$. The second scenario yields $\log_2\lceil |A| + 1 \rceil$ propositional variables - all the possible combinations of Boolean values are allowed as assignments while all the propositional variables set to $FALSE$ correspond to no occupancy of the location. If the preference of assigning $FALSE$ actually results in setting strictly fewer variables to $TRUE$ then no occupancy immediately appears in the first scenario while there is little chance that no occupancy appears in the second scenario (setting strictly fewer variables to $TRUE$ may lead to another assignment of the bit vector with some propositional variables set to $TRUE$ - that is, representing some occupancy).

Results of measurement of the total number of moves generated by SAT based CPF solving with suggested encodings are presented in Figure 10, Figure 11, and Figure 12. The same set of testing instances over 4-connected grids as in the runtime measurement has been used. The total number of moves generated by OD+ID is also included in the measurement.

The fewest number of moves in most testing instances is yielded by the SIMPLIFIED encoding. Thus, we also present sorted differences in the total number of moves between those yielded by the SIMPLIFIED encoding and other methods. It can be also observed that OD+ID generates solutions with the smallest number of moves in instances containing few agents. Unfortunately, as OD+ID does not scale up well enough it cannot show qualities of its solutions for larger number of agents.

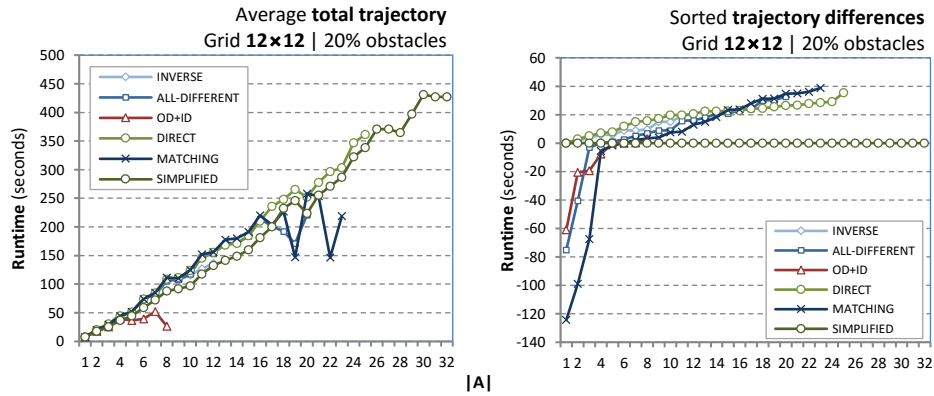

**Figure 12.** *Solution quality comparison over 12 × 12 4-connected grid.* The SIMPLIFIED encoding yields almost always a solution with fewer moves than other methods in this larger scenario. Again, solutions with fewest moves are generated by OD+ID but the comparison could be done for few agents only due to insufficient scalability of OD+ID.

There is almost no significant difference in the total number of moves generated by other methods except a marginal tendency of the DIRECT encoding to yield better solutions than methods using binary encoded bit vectors especially observable over 8 × 8 grid.

Altogether, we can conclude that the hypothesis that encodings using directly encoded states is more advantageous than encodings with binary encoded bit-vectors with respect to the total number of moves.

## 6. Conclusions

Several propositional encodings of cooperative path-finding problem (CPF) have been introduced - INVERSE, ALL-DIFFERENT, MATCHING, DIRECT, and SIMPLIFIED encodings. The presented encodings are based on the notion of the time expanded graph that expands the graph modeling the environment over time so that arrangements of agents at all the time steps up to a certain final time step can be represented. Time expanded graphs provided an essential step towards building propositional formulae, in which a query whether there is a solution of a given CPF with the specified number of time steps is encoded. Obtaining makespan optimal solution is then carried out by submitting multiple encoded queries to a SAT solver. The reduction of CPF to SAT allows accessing all the advanced search, pruning, and learning techniques of the SAT solver that can be in this way employed in CPF solving.

The suggested encodings either use binary encoded bit vectors (INVERSE, ALL-DIFFERENT, and MATCHING encoding) or directly encoded states (DIRECT and SIMPLIFIED encodings) to model arrangements of agents at individual time steps. Using binary encoded bit vectors results in smaller formulae in terms of the number of variables and clauses. The advantage of encodings with directly encoded states is on the other hand a better support for Boolean constraint propagation (unit propagation) which enabled by the presence of many short clauses.

Performed experimental evaluation indicates that CPF solving via SAT is generally the best option in highly constrained situations (environments densely occupied by agents). SAT based CPF solving scales up for larger number of agents much better than the alternative A* based search technique.

If we compare solely SAT encodings, than the SIMPLIFIED encoding turned out to perform as best. Instances with the highest occupancy by agents were solved only by the SIMPLIFIED encoding in the given timeout. Moreover, the comparison of the quality of solutions generated by the SAT based solving in terms of the total number of generated moves also indicates that the SIMPLIFIED encoding generates fewest moves.

**Acknowledgments**

This work is supported by Czech-Israeli cooperation project number 8G15027 and by Charles University within the PRVOUK (section P46) and UNCE projects. The author would like to thank the research team from Ben Gurion University, Israel led by Ariel Felner and Roni Stern and student members of the team Guni Sharon and Eli Boyarski for providing source code for conducting presented experiments.

*Makespan Optimal Solving of Cooperative Path-Finding* 396. **Biere**, A., **Brummayer**, R. *Consistency Checking of All Different Constraints over Bit-Vectors within a SAT Solver.* Proceedings of Formal Methods in Computer-Aided Design (FMCAD 2008), pp. 1-4, IEEE Press, 2008.
7. **Biere**, A., **Heule**, M., **van Maaren**, H., **Walsh**, T. *Handbook of Satisfiability.* IOS Press, 2009.
8. **Čáp**, M., **Novák**, P., **Vokřínek**, J., **Pěchouček**, M. *Multi-agent RRT: sampling-based cooperative pathfinding.* International conference on Autonomous Agents and Multi-Agent Systems (AAMAS 2013), pp. 1263-1264, IFAAMAS, 2013.
9. **Erdem**, E., **Kisa**, D. G., **Öztok**, U., **Schüller**, P. *A General Formal Framework for Pathfinding Problems with Multiple Agents.* Proceedings of the 27th AAAI Conference on Artificial Intelligence (AAAI 2013), AAAI Press, 2013.
10. **Huang**, R., **Chen**, Y., **Zhang**, W. *A Novel Transition Based Encoding Scheme for Planning as Satisfiability.* Proceedings of the 24th AAAI Conference on Artificial Intelligence (AAAI 2010), AAAI Press, 2010.
11. **Kautz**, H., **Selman**, B. *Unifying SAT-based and Graph-based Planning.* Proceedings of the 16th International Joint Conference on Artificial Intelligence (IJCAI 1999), pp. 318-325, Morgan Kaufmann, 1999.
12. **Kim**, D., **Hirayama**, K., **Park**, G.-K. *Collision Avoidance in Multiple-Ship Situations by Distributed Local Search.* Journal of Advanced Computational Intelligence and Intelligent Informatics (JACIII), Volume 18(5), pp. 839-848, Fujipress, 2014.
13. **KIVA** Systems. *Official web site.* http://www.kivasystems.com/, 2015 [accessed in April 2015].
14. **Kornhauser**, D., **Miller**, G. L., and **Spirakis**, P. G.: *Coordinating Pebble Motion on Graphs, the Diameter of Permutation Groups, and Applications.* Proceedings of the 25th Annual Symposium on Foundations of Computer Science (FOCS 1984), pp. 241-250, IEEE Press, 1984.
15. **Michael**, N., **Fink**, J., **Kumar**, V. Cooperative manipulation and transportation with aerial robots. Autonomous Robots, Volume 30(1), pp. 73-86, Springer, 2011.
16. **Ratner**, D., **Warmuth**, M. K.: *Finding a Shortest Solution for the N × N Extension of the 15-PUZZLE Is Intractable.* Proceedings of the 5th National Conference on Artificial Intelligence (AAAI 1986), pp. 168-172, Morgan Kaufmann, 1986.
17. **Rintanen**, J., **Heljanko**, K., and **Niemelä**, I. *Planning as satisfiability: parallel plans and algorithms for plan search.* Artificial Intelligence, Volume 170 (12-13), pp. 1031–1080, Elsevier, 2006.
18. **Ryan**, M. R. K. *Graph Decomlocation for Efficient Multi-Robot Path Planning.* Proceedings of the 20th International Joint Conference on Artificial Intelligence (IJCAI 2007), pp. 2003-2008, IJCAI Conference, 2007.
19. **Ryan**, M. R. K. *Exploiting Subgraph Structure in Multi-Robot Path Planning.* Journal of Artificial Intelligence Research (JAIR), Volume 31, pp. 497-542, AAAI Press, 2008.
20. **Silver**, D.: *Cooperative Pathfinding.* Proceedings of the 1st Artificial Intelligence and Interactive Digital Entertainment Conference (AIIDE 2005), pp. 117-122, AAAI Press, 2005.
21. **Sharon**, G., **Stern**, R., **Goldenberg**, M., **Felner**, A.: *The increasing cost tree search for optimal multi-agent pathfinding.* Artificial Intelligence, Volume 195, pp. 470-495, Elsevier, 2013.
22. **Sharon**, G., **Stern**, R., **Felner**, A., **Sturtevant**, N. R.: *Conflict-based search for optimal multi-agent pathfinding.* Artificial Intelligence, Volume 219, pp. 40-66, Elsevier, 2015.
23. **Standley**, T. S., **Korf**, R. E.: *Complete Algorithms for Cooperative Pathfinding Problems.* Proceedings of Proceedings of the 22nd International Joint Conference on Artificial Intelligence (IJCAI 2011), pp. 668-673, IJCAI/AAAI Press, 2011.
24. **Sturtevant**, N. R. *Benchmarks for Grid-Based Pathfinding.* IEEE Transactions on Computational Intelligence and AI in Games, Volume 4(2), pp. 144-148, IEEE Press, 2012.